\documentclass[runningheads]{llncs}
\usepackage{graphicx}

\usepackage[pagebackref,breaklinks,colorlinks]{hyperref}

\usepackage{tikz}
\usepackage{comment}
\usepackage{amsmath,amssymb} 

\usepackage{mmstyle}

\usepackage{cite}
\usepackage{booktabs} 
\usepackage{color}
\usepackage{xspace}
\usepackage{ctable}
\usepackage{color}
\usepackage{subcaption}
\usepackage{multirow}
\usepackage{float}
\usepackage{stfloats}
\usepackage{stmaryrd}
\usepackage{algorithm}
\usepackage{algorithmicx}
\usepackage{algpseudocode}
\usepackage{threeparttable}
\usepackage{wrapfig}
\usepackage{orcidlink}

\usepackage[accsupp]{axessibility}  

\newcommand{\name}{HuMMan\xspace}
\newcommand{\numsubjects}{1000\xspace}
\newcommand{\numactions}{500\xspace}
\newcommand{\numsequences}{400k\xspace}
\newcommand{\numframes}{60M\xspace}

\newcommand{\Fig}{Fig.\xspace}
\newcommand{\Tab}{Table\xspace}
\newcommand{\Sec}{Section\xspace}
\newcommand{\Supp}{Supplementary Material\xspace}

\usepackage{xcolor}
\usepackage{color, colortbl}
\definecolor{Gray}{gray}{0.9}

\newcommand{\repeatthanks}{\textsuperscript{\thefootnote}}

\begin{document}
\pagestyle{headings}
\mainmatter
\def\ECCVSubNumber{185}  

\title{HuMMan: Multi-Modal 4D Human Dataset for Versatile Sensing and Modeling}

\author{
Zhongang Cai\thanks{co-first authors; $^{\dagger}$ co-corresponding authors}$^{,1,2,3}$\orcidlink{0000-0002-1810-3855}, 
Daxuan Ren\repeatthanks$^{,2}$\orcidlink{0000-0002-8449-3038}, 
Ailing Zeng\repeatthanks$^{,4}$\orcidlink{0000-0002-3783-0679}, 
Zhengyu Lin\repeatthanks$^{,3}$\orcidlink{0000-0003-4173-9752},
Tao Yu\repeatthanks$^{,5}$\orcidlink{0000-0002-3818-5069}, 
Wenjia Wang$\repeatthanks^{,3}$\orcidlink{0000-0003-0121-3852},
Xiangyu Fan$^{3}$\orcidlink{0000-0002-3446-524X}, 
Yang Gao$^{3}$\orcidlink{0000-0001-6505-7081}, 
Yifan Yu$^{3}$\orcidlink{0000-0002-5290-8278}, 
Liang Pan$^{2}$\orcidlink{0000-0003-1821-4296}, 
Fangzhou Hong$^{2}$\orcidlink{0000-0003-2412-1141}, 
Mingyuan Zhang$^{2}$\orcidlink{0000-0001-8212-715X}, 
Chen Change Loy$^{2}$\orcidlink{0000-0001-5345-1591}\index{Loy, Chen Change},
Lei Yang$^{\dagger,1,3}$\orcidlink{0000-0002-0571-5924}, 
Ziwei Liu$^{\dagger,2}$\orcidlink{0000-0002-4220-5958}
}

\institute{$^{1}$Shanghai AI Laboratory, $^{2}$S-Lab, Nanyang Technological University, $^{3}$SenseTime Research, $^{4}$The Chinese University of Hong Kong, $^{5}$Tsinghua University
\email{yanglei@sensetime.com, ziwei.liu@ntu.edu.sg}
}

\titlerunning{HuMMan}
\authorrunning{Z. Cai et al.}

\maketitle

\begin{abstract}
    4D human sensing and modeling are fundamental tasks in vision and graphics with numerous applications. With the advances of new sensors and algorithms, there is an increasing demand for more versatile datasets. In this work, we contribute \textbf{\name}, a large-scale multi-modal 4D human dataset with \numsubjects human subjects, \numsequences sequences and \numframes frames. \name has several appealing properties: \textbf{1)} multi-modal data and annotations including color images, point clouds, keypoints, SMPL parameters, and textured meshes; \textbf{2)} popular mobile device is included in the sensor suite; \textbf{3)} a set of 500 actions, designed to cover fundamental movements; \textbf{4)} multiple tasks such as action recognition, pose estimation, parametric human recovery, and textured mesh reconstruction are supported and evaluated. Extensive experiments on \name voice the need for further study on challenges such as fine-grained action recognition, dynamic human mesh reconstruction, point cloud-based parametric human recovery, and cross-device domain gaps.\footnote{Homepage: \url{https://caizhongang.github.io/projects/HuMMan/}}
\end{abstract}


\section{Introduction}
\begin{figure}[t]
  \includegraphics[width=\linewidth]{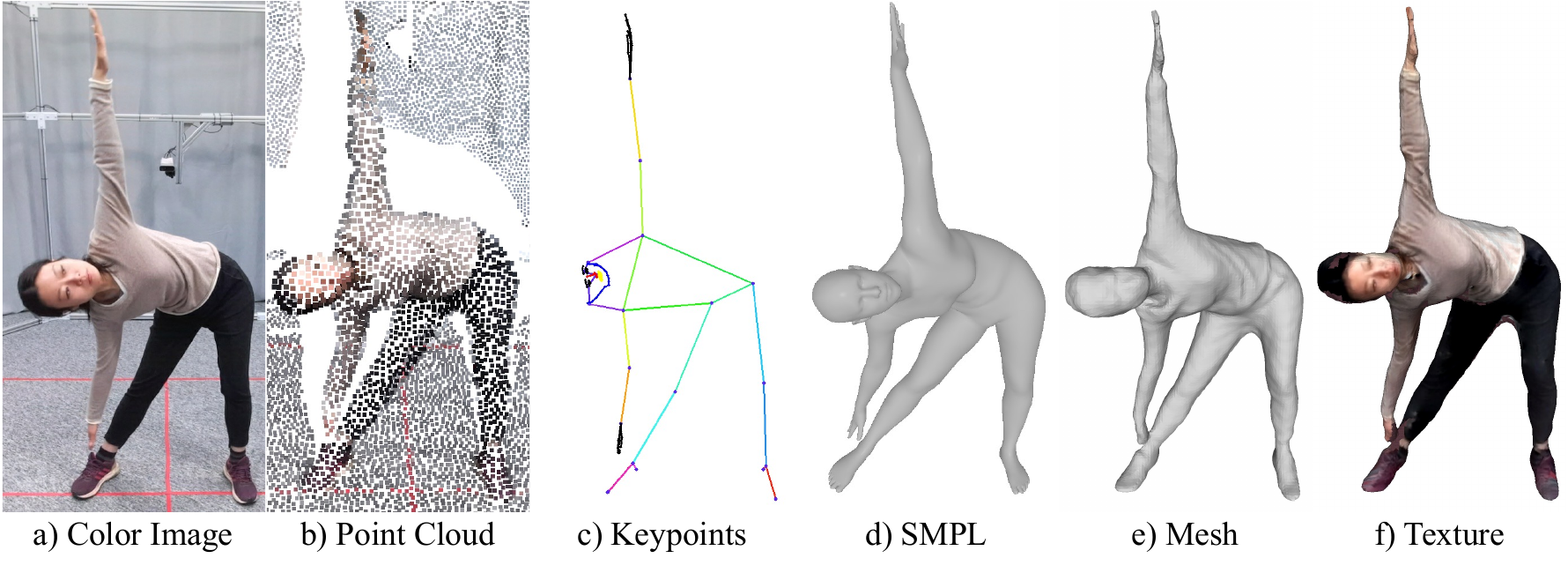}
  \caption{\name features multiple modalities of data format and annotations. We demonstrate a) color image, b) point cloud,  c) keypoints, d) SMPL parameters and e) mesh geometry with f) texture. Each sequence is also annotated with an action label from 500 actions. Each subject has two additional high-resolution scans of naturally and minimally clothed body. }
  \label{fig:modality}
\end{figure}

Sensing and modeling humans are longstanding problems for both computer vision and computer graphics research communities, which serve as the fundamental technology for a myriad of applications such as animation, gaming, augmented, and virtual reality. With the advent of deep learning, significant progress has been made alongside the introduction of large-scale datasets in human-centric sensing and modeling \cite{ionescu2013human3, lin2014microsoft, von2018recovering, ma2020learning, Zheng2019DeepHuman, Yu2020HUMBIAL}. In this work, we present \textbf{\name}, a comprehensive human dataset consisting of \numsubjects human subjects, captured in total \numsequences sequences and \numframes frames. More importantly, \name features four main properties listed below. 

\textbf{- Multiple Modalities}. 
\name provides a basket of data formats and annotations in the hope to assist exploration in their potential complementary nature. We build \name with a set of 10 synchronized RGB-D cameras to capture both video and depth sequences. Our toolchain then post-process the raw data into sequences of colored point clouds, 2D/3D keypoints, statistical model (SMPL) parameters, and model-free textured mesh. Note that all data and annotations are temporally synchronized, while 3D data and annotations are spatially aligned. In addition, we provide a high-resolution scan for each of the subjects in a canonical pose. 


\textbf{- Mobile Device}.
With the development of 3D sensors, it is common to find depth cameras or low-power LiDARs on a mobile device in recent years. In view of the surprising gap between emerging real-life applications and the insufficiency of data collected with mobile devices, we add a mobile phone with built-in LiDAR in the data collection to facilitate the relevant research.

\textbf{- Action Set}. 
We design \name to empower comprehensive studies on human actions. Instead of empirically selecting daily activities, we propose to take an anatomical point of view and systematically divide body movements by their driving muscles. Specifically, we design \numactions movements by categorizing major muscle groups to achieve a more complete and fundamental representation of human actions. 

\textbf{- Multiple Tasks}. 
To facilitate research on \name, we provide a whole suite of baselines and benchmarks for action recognition, 2D and 3D pose estimation, 3D parametric human recovery, and textured mesh reconstruction. Popular methods are implemented and evaluated using standard metrics. Our experiments demonstrate that \name would be useful for multiple fields of study, such as fine-grained action recognition, point cloud-based parametric human recovery, dynamic mesh sequence reconstruction, and transferring knowledge across devices.

In summary, \name is a large-scale multi-modal dataset for 4D (spatio-temporal) human sensing and modeling, with four main features: \textbf{1)} multi-modal data and annotations; \textbf{2)} mobile device included in the sensor suite; \textbf{3)} action set with atomic motions; \textbf{4)} standard benchmarks for multiple vision tasks. We hope \name would pave the way towards more comprehensive sensing and modeling of humans.



\section{Related Works}
\label{sec:related_works}
\begin{table}[t]
\caption{
Comparisons of \name with published datasets. \name has a competitive scale in terms of the number of subjects (\#Subj), actions (\#Act), sequences (\#Seq) and frames (\#Frame). Moreover, \name features multiple modalities and supports multiple tasks. Video: sequential data, not limited to RGB sequences; Mobile: mobile device in the sensor suite; D/PC: depth image or point cloud, only genuine point cloud collected from depth sensors are considered; Act: action label; K2D: 2D keypoints; K3D: 3D keypoints; Param: statistical model (\eg SMPL) parameters; Txtr: texture. -: not applicable or not reported.
}
\label{tab:action_set_comparison}
\centering
\scriptsize
\resizebox{\textwidth}{!}{
\begin{tabular}{lcccccccccccccc}

  \toprule

  \multirow{2}{*}{Dataset} & 
  \multirow{2}{*}{\#Subj} & 
  \multirow{2}{*}{\#Act} & 
  \multirow{2}{*}{\#Seq} & 
  \multirow{2}{*}{\#Frame} & 
  \multirow{2}{*}{Video} & 
  \multirow{2}{*}{Mobile} & 
  \multicolumn{8}{c}{Modalities} \\
  \cmidrule{8-15}
  &&&&&&& RGB & D/PC & Act & K2D & K3D & Param & Mesh & Txtr \\
  
  
  \midrule

  
  UCF101~\cite{soomro2012ucf101} & 
  - & 101 & 13k & - & \checkmark & - & 
  \checkmark & - & \checkmark & - & - & - & - & - \\
  

  AVA~\cite{gu2018ava} & 
  - & 80 & 437 & - & \checkmark & - & 
  \checkmark & - & \checkmark & - & - & - & - & - \\
  


  
  FineGym~\cite{shao2020finegym} & 
  - & 530 & 32k & - & \checkmark & - & 
  \checkmark & - & \checkmark & - & - & - & - & - \\

  HAA500~\cite{chung2021haa500} & 
  - & 500 & 10k & 591k & \checkmark & - & 
  \checkmark & - & \checkmark & - & - & - & - & - \\

  
  
  SYSU 3DHOI~\cite{hu2015jointly} & 
  40 & 12 & 480 & - & \checkmark & - & 
  \checkmark & \checkmark & \checkmark & - & \checkmark & - & - & - \\
  
  NTU RGB+D~\cite{shahroudy2016ntu} & 
  40 & 60 & 56k & - & \checkmark & - & 
  \checkmark & \checkmark & \checkmark & - & \checkmark & - & - & - \\
  
  NTU RGB+D 120~\cite{liu2019ntu} & 
  106 & 120 & 114k & - & \checkmark & - & 
  \checkmark & \checkmark & \checkmark & - & \checkmark & - & - & - \\
  
  NTU RGB+D X~\cite{trivedi2021ntu} & 
  106 & 120 & 113k & - & \checkmark & - & 
  \checkmark & \checkmark & \checkmark & - & \checkmark & \checkmark & - & - \\

  \midrule
  
    
  
  MPII~\cite{andriluka20142d} & 
  - & 410 & - & 24k & - & - & 
  \checkmark & - & \checkmark & \checkmark & - & - & - & - \\
  
  COCO~\cite{lin2014microsoft} & 
  - & - & - & 104k & - & - & 
  \checkmark & - & - & \checkmark & - & - & - & - \\
  
  PoseTrack~\cite{andriluka2018posetrack} & 
  - & - & $>$1.35k & $>$46k & \checkmark & - & 
  \checkmark & - & - & \checkmark & - & - & - & - \\  

  Human3.6M~\cite{ionescu2013human3} & 
  11 & 17 & 839 & 3.6M & \checkmark & - & 
  \checkmark & \checkmark & \checkmark & \checkmark & \checkmark & - & - & - \\ 
  
  CMU Panoptic~\cite{joo2015panoptic} & 
  8 & 5 & 65 & 154M & \checkmark & - & 
  \checkmark & \checkmark & - & \checkmark & \checkmark & - & - & - \\
  
  MPI-INF-3DHP~\cite{mehta2017monocular} & 
  8 & 8 & 16 & 1.3M & \checkmark & - & 
  \checkmark & - & - & \checkmark & \checkmark & - & - & - \\
  
      
  3DPW~\cite{von2018recovering} & 
  7 & - & 60 & 51k & \checkmark & \checkmark & 
  \checkmark & - & - & - & - & \checkmark & - & - \\ 
      
  AMASS~\cite{mahmood2019amass} & 
  344 & - & $>$11k & $>$16.88M & \checkmark & - & 
  - & - & - & - & \checkmark & \checkmark & - & - \\ 


  AIST++~\cite{li2021ai} & 
  30 & - & 1.40k & 10.1M & \checkmark & - & 
  \checkmark & - & - & \checkmark & \checkmark & \checkmark & - & - \\

  \midrule

  CAPE~\cite{ma2020learning} & 
  15 & - & $>$600 & $>$140k & \checkmark & - & 
  - & - & \checkmark & - & \checkmark & \checkmark & \checkmark & - \\  

  BUFF~\cite{zhang2017detailed} & 
  6 & 3 & $>$30 & $>$13.6k & \checkmark & - & 
  \checkmark & \checkmark & \checkmark & - & \checkmark & \checkmark & \checkmark & \checkmark \\

  DFAUST~\cite{bogo2017dynamic} & 
  10 & $>$10 & $>$100 & $>$40k & \checkmark & - & 
  \checkmark & \checkmark & \checkmark & \checkmark & \checkmark & \checkmark & \checkmark & \checkmark \\  





  HUMBI~\cite{Yu2020HUMBIAL} & 
  772 & - & - & $\sim$26M & \checkmark & - & 
  \checkmark & - & - & \checkmark & \checkmark & \checkmark & \checkmark & \checkmark \\
  
  ZJU LightStage~\cite{peng2021neural} & 
  6 & 6 & 9 & $>$1k & \checkmark & - & 
  \checkmark & - & \checkmark & \checkmark & \checkmark & \checkmark & \checkmark & \checkmark \\


  THuman2.0~\cite{tao2021function4d} & 
  200 & - & - & $>$500 & - & - & 
  - & - & - & - & - & \checkmark & \checkmark & \checkmark \\

  \midrule
  
  \textbf{\name (ours)}  & \numsubjects & \numactions & \numsequences & \numframes & \checkmark & \checkmark & \checkmark & \checkmark & \checkmark & \checkmark & \checkmark & \checkmark & \checkmark & \checkmark \\
  
  \bottomrule

\end{tabular}
}
\end{table}

\noindent \textbf{Action Recognition.} As an important step towards understanding human activities, action recognition is the task to categorize human motions into predefined classes. RGB videos~\cite{tran2018closer, feichtenhofer2019slowfast, tran2019video, feichtenhofer2020x3d} with additional information such as optical flow and estimated poses and 3D skeletons typically obtained from RGB-D sequences~\cite{yan2018spatial, shi2019skeleton, shi2020skeleton, zeng2021learning} are the common input to existing methods.
%
%
Datasets for RGB video-based action recognition are often collected from the Internet. Some have a human-centric action design~\cite{kuehne2011hmdb, soomro2012ucf101, karpathy2014large, gu2018ava, shao2020finegym, chung2021haa500} whereas others introduce interaction and diversity in the setup~\cite{carreira2019short, monfort2019moments, zhao2019hacs}. Recently, fine-grained action understanding~\cite{gu2018ava, shao2020finegym, chung2021haa500} is drawing more research attention. However, these 2D datasets lack 3D annotations. As for RGB-D datasets, earlier works are small in scale~\cite{li2010action, wang2014cross, hu2015jointly}. As a remedy, the latest NTU RGB-D series~\cite{liu2019ntu, shahroudy2016ntu, trivedi2021ntu} features 60-120 actions. However, the majority of the actions are focused on the upper body. We develop a larger and more complete action set in \name.



\noindent \textbf{2D and 3D Keypoint Detection.} Estimation of a human pose is a vital task in computer vision, and a popular pose representation is human skeletal keypoints. The field is categorized by output format: 2D~\cite{newell2016stacked, chen2018cascaded, sun2019hrnet, li2021rle} and 3D~\cite{martinez2017simple, zhao2019semantic, pavllo20193d, zeng2020srnet, zeng2021learning, zeng2021smoothnet} keypoint detection, or by the number of views: single-view~\cite{newell2016stacked, chen2018cascaded, sun2019hrnet, pavllo20193d, martinez2017simple, zhao2019semantic, zeng2021learning} and multi-view pose estimation~\cite{qiu2019cross, iskakov2019learnable, huang2019deepfuse}.
For 2D keypoint detection, single-frame datasets such as MPII~\cite{andriluka20142d} and COCO~\cite{lin2014microsoft} provide diverse images with 2D keypoints annotations, whereas video datasets such as J-HMDB~\cite{jhuang2013towards}, Penn Action~\cite{zhang2013actemes} and PoseTrack~\cite{andriluka2018posetrack} provide sequences of 2D keypoints. However, they lack 3D ground truths. In contrast, 3D keypoint datasets are typically built indoor data to accommodate sophisticated equipment, such as Human3.6M~\cite{ionescu2013human3}, CMU Panoptic~\cite{joo2015panoptic}, MPI-INF-3DHP~\cite{mehta2017monocular}, TotalCapture~\cite{trumble2017total}, and AIST++~\cite{li2021ai}. Compared to these datasets, \name not only supports 2D and 3D keypoint detection but also textured mesh reconstruction assist in more holistic modeling of humans.

\noindent \textbf{3D Parametric Human Recovery.}
Also known as human pose and shape estimation, 3D parametric human recovery leverages human parametric model representation (such as SMPL \cite{loper2015smpl}, SMPL-X \cite{pavlakos2019expressive}, STAR \cite{osman2020star} and GHUM \cite{xu2020ghum}) that achieves sophisticated mesh reconstruction with a small amount of parameters. Existing methods take keypoints \cite{bogo2016keep, pavlakos2019expressive, Zhang2021LightWeight}, images \cite{pavlakos2018learning, omran2018neural, kolotouros2019learning, guler2019holopose, georgakis2020hierarchical, li2020hybrik, kocabas2021pare}, videos \cite{kanazawa2019learning, sun2019human, mehta2020xnect, moon2020i2l, choi2021beyond, luo20203d}, and point clouds \cite{jiang2019skeleton, bhatnagar2020combining, wang2021locally, liu2021votehmr} as the input to obtain the parameters. Joint limits \cite{akhter2015pose} and contact \cite{muller2021self} are also important research topics.
Apart from those that provide keypoints, various datasets also provide ground-truth SMPL parameters. MoSh \cite{loper2014mosh} is applied on Human3.6M \cite{ionescu2013human3} to generate SMPL annotations. CMU Panoptic \cite{joo2015panoptic} and HUMBI \cite{Yu2020HUMBIAL} leverages keypoints from multiple camera views. 3DPW \cite{von2018recovering} combines a mobile phone and inertial measurement units (IMUs). Synthetic dataset such as AGORA \cite{patel2021agora} renders high-quality human scans in virtual environments and fits SMPL to the original mesh. Video games have also become an alternative source of data \cite{cao2020long, cai2021playing}. In addition to SMPL parameters that do not model clothes or texture, \name also provides textured meshes of clothed subjects. 

\noindent \textbf{Textured Mesh Reconstruction.}
%
%
To reconstruct the 3D surface, common methods include multi-view stereo~\cite{Furukawa2007PMVS}, volumetric fusion~\cite{izadi2011kinectfusion, newcombe2015dynamicfusion, yu2018doublefusion}, Poisson surface reconstruction~\cite{kazdan2013Poisson, Kazhdan2020PSREC}, and neural surface reconstruction~\cite{Peng2020ECCV_Full, saito2019pifu}. To reconstruct texture for the human body, popular approaches include texture mapping or montage~\cite{Gal2010TextureMontage}, deep neural rendering~\cite{Lombardi2018DeepAppearance}, deferred neural rendering~\cite{Thies2019DNR}, and NeRF-like methods~\cite{mildenhall2020nerf}.
%
%
%
Unfortunately, existing datasets for textured human mesh reconstruction typically provide no sequential data~\cite{Zheng2019DeepHuman, tao2021function4d}, which is valuable to the reconstruction of animatable avatars~\cite{Xiang2021fullbodyavatar,raj2020anr}. Moreover, many have only a limited number of subjects~\cite{peng2021neural, ma2020learning, zhang2017detailed, bogo2017dynamic, alldieck2018video, habermann2019livecap, habermann2021real, habermann2020deepcap}. In contrast, \name includes diverse subjects with high-resolution body scans and a large amount of dynamic 3D sequences.
\section{Hardware Setup}
\label{sec:hardware}
\begin{figure}[t]
  \includegraphics[width=\linewidth]{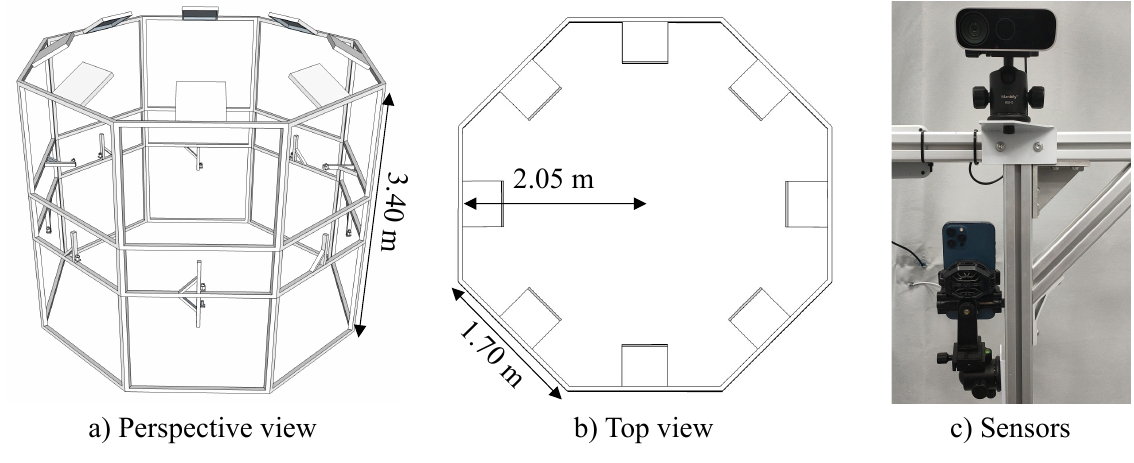}
  \caption{Hardware setup. a) and b) we build a octagonal prism-shaped framework to accommodate the data collection system. c) sensors used to collect sequential data include ten Azure Kinects and an iPhone 12 Pro Max. Besides, an Artec Eva is used to produce high-resolution static scans of the subjects.}
  \label{fig:hardware}
\end{figure}

We customize an octagonal prism-shaped multi-layer framework to accommodate calibrated and synchronized sensors. The system is 1.7 m in height and 3.4 m in side length of its octagonal cross-section as illustrated in \Fig\ref{fig:hardware}.

\subsection{Sensors}

\noindent \textbf{RGB-D Sensors.} 
Azure Kinect is popular with both academia and the industry with a color resolution of 1920$\times$1080, and a depth resolution of 640$\times$576. We deploy ten Kinects to capture multi-view RGB-D sequences. The Kinects are strategically placed to ensure a uniform spacing, and a wide coverage such that any body part of the subject, even in most expressive poses, is visible to at least two sensors. We develop a program that interfaces with Kinect's SDK to obtain a data throughput of 74.4 MB per frame and 2.2 GB per second at 30 FPS before data compression. 

\noindent \textbf{Mobile Device.} 
An iPhone 12 Pro Max is included in the sensor suite to allow for the study on a mobile device. Besides the regular color images of resolution 1920$\times$1440, the built-in LiDAR produces depth maps of resolution 256$\times$192. We develop an iOS app upon ARKit to retrieve the data.

\noindent \textbf{High-Resolution Scanner.} 
To supplement our sequential data with high-quality body shape information, a professional handheld 3D scanner, Artec Eva, is used to produce a body scan of resolution up to 0.2 mm and accuracy up to 0.1 mm. A typical scan consists of $300k$ to $500k$ faces and $100k$ to $300k$ vertices, with a 4K (4096$\times$4096) resolution texture map. 

\subsection{Two-Stage Calibration}

\noindent \textbf{Image-based Calibration.}
To obtain a coarse calibration, we first perform image-based calibration following the general steps in Zhang's method~\cite{zhang2000flexible}. However, we highlight that Kinect's active IR depth cameras encounter over-exposure with regular chessboards. Hence, we customize a light absorbent material to cover the black squares of the chessboard pattern. In this way, we acquire reasonably accurate extrinsic calibration for Kinects and iPhones.

\noindent \textbf{Geometry-based Calibration.}
Image-based calibration is unfortunately not accurate enough to reconstruct good-quality mesh. Hence, we propose to take advantage of the depth information in a geometry-based calibration stage. We empirically verify that image-based calibration serves as a good initialization for geometry-based calibration. Hence, we randomly place stacked cubes inside the framework. After that, we convert captured depth maps to point clouds and apply multi-way ICP registration~\cite{choi2015robust} to refine the calibration. 

\subsection{Synchronization}
\noindent \textbf{Kinects.} As the Azure Kinect implements the Time-of-Flight principle, it actively illuminates the scene multiple times (nine exposures in our system) for depth computation. To avoid interference between individual sensors, we use the synchronization cables to propagate a unified clock in a daisy chain fashion, and reject any image that is 33 ms or above out of synchronization. We highlight that there is only a 1450-us interval between exposures of 160 us; our system of ten Kinects reaches the theoretical maximum number. 

\noindent \textbf{Kinect-iPhone.} Due to hardware limitations, we cannot apply the synchronization cable to the iPhone. We circumvent this challenge by implementing a TCP-based communication protocol that computes an offset between the Kinect clock and the iPhone ARKit clock. As iPhone is recording at 60 FPS, we then use the offset to map the closest iPhone frames to Kinect frames. Our test shows the synchronization error is constrained below 33 ms.

\section{Toolchain}
\label{sec:toolchain}
\begin{figure}[t]
  \includegraphics[width=\linewidth]{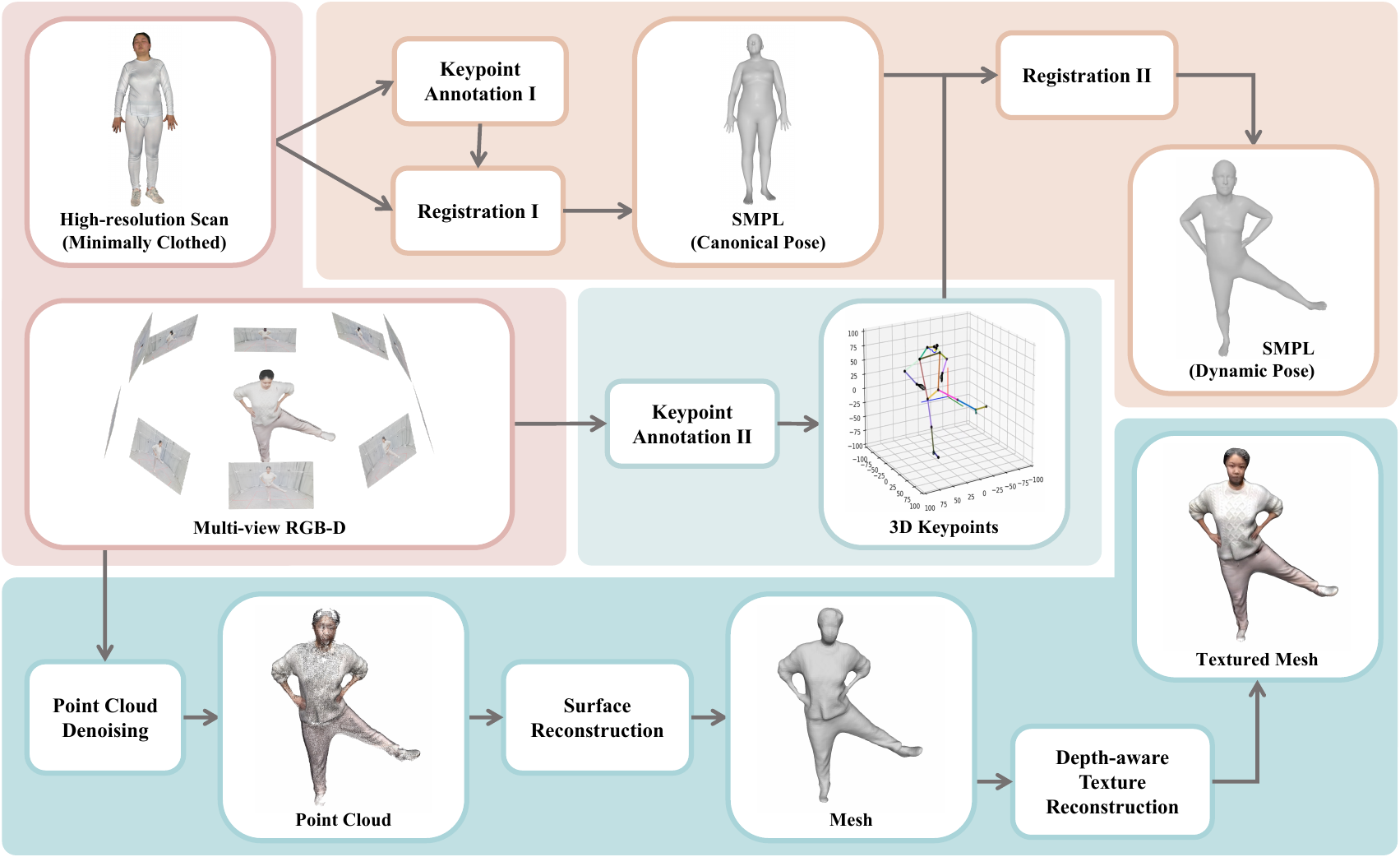}
  \caption{Our toolchain produces multiple annotation formats such as 3D keypoint sequences, SMPL parameter sequences, and textured mesh sequences}
  \label{fig:toolchain}
\end{figure}

To handle the large volume of data, we develop an automatic toolchain to provide annotations such as keypoints and SMPL parameters. Moreover, dynamic sequences of textured mesh are also reconstructed. The pipeline is illustrated in \Fig\ref{fig:toolchain}. Note that there is a human inspection stage to reject low-quality data with erroneous annotations.

\subsection{Keypoint Annotation}

There are two stages of keypoint annotation (I and II) in the toolchain. For stage I, virtual cameras are placed around the minimally clothed body scan to render multi-view images. For stage II, the color images from multi-view RGB-D are used. The core ideas of the keypoint annotation are demonstrated below, with the detailed algorithm in the \Supp.

\noindent \textbf{Multi-view 2D Keypoint Detection.}
We employ the whole-body pose model that includes body, hand and face 2D keypoints $\hat{\cP}_{2D} \in \mathbb{R}^{P \times 2}$, where $P=133$. A large deep learning model HRNet-w48~\cite{sun2019hrnet} is used which achieves AP 66.1 and AR 74.3 on COCO whole-body benchmark~\cite{jin2020whole}.

\noindent \textbf{3D Keypoint Triangulation.}
As the camera intrinsic and extrinsic parameters are available, we triangulate 3D keypoints $\cP_{3D} \in \mathbb{R}^{P \times 3}$ with the multi-view 2D estimated keypoints $\hat{\cP}_{2D}$. However, 2D keypoints from any single view may not be always reliable. Hence, we use the following strategies to improve the quality of 3D keypoints.
1) \textit{Keypoint selection}. To avoid the influence of poor-quality estimated 2D keypoints, we use a threshold $\tau_k$ to remove keypoints with a low confidence score.
2) \textit{Camera selection}. As our system consists of ten Kinects, we exploit the redundancy to remove low-quality views. We only keep camera views with reprojection errors that are top-$k$ smallest~\cite{karashchuk2021anipose} and no larger than a threshold $\tau_c$.
3) \textit{Smoothness constraint}. Due to inevitable occlusion in the single view, the estimated 2D keypoints often have jitters. To alleviate the issue, we develop a smoothness loss to minimize the difference between consecutive triangulated 3D keypoints. Note that we design the loss weight to be inversely proportional to average speed, in order to remove jitters without compromising the ability to capture fast body motions.
4) \textit{Bone length constraint}. As human bone length is constant, the per-frame bone length is constrained towards the median bone length $\cB$ pre-computed from the initial triangulated 3D keypoints.
The constraints are formulated as Eq.~\ref{eq:tri}:
%
\begin{equation}
    E_{tri} = \lambda_1\sum_{t=0}^{T-1} \|\cP_{3D}(t+1) - \cP_{3D}(t)\| + \lambda_2\sum_{(i, j) \in \cI_{\cB}} \|\cB_{i, j} - f_{\cB}(\cP_{3D}(i, j)) \|
    \label{eq:tri}
\end{equation}
where $\cI_{\cB}$ contains the indices of connected keypoints and $f_{\cB}(\cdot)$ calculates the average bone length of a given 3D keypoint sequence. Note that 3) and 4) are jointly optimized.

\noindent \textbf{2D Keypoint Projection.}
To obtain high-quality 2D keypoints $\cP_{2D} \in \mathbb{R}^{P \times 2}$, we project the triangulated 3D keypoints to image space via calibrated camera parameters. Note that this step is only needed for stage II keypoint annotation.

\noindent \textbf{Keypoint Quality.}
We use $\cP_{2D}$ and $\cP_{3D}$ as keypoint annotations for 2D Pose Estimation and 3D Pose Estimation, respectively. To gauge the accuracy of the automatic keypoint annotation pipeline, we manually annotate a subset of data. The average Euclidean distance between annotated 2D keypoints and reprojected 2D keypoints $\cP_{2D}$ is $15.13$ pixels on the resolution of $1920\times1080$.



\subsection{Human Parametric Model Registration}
\begin{figure}[t]
    \centering
    \includegraphics[width=\linewidth]{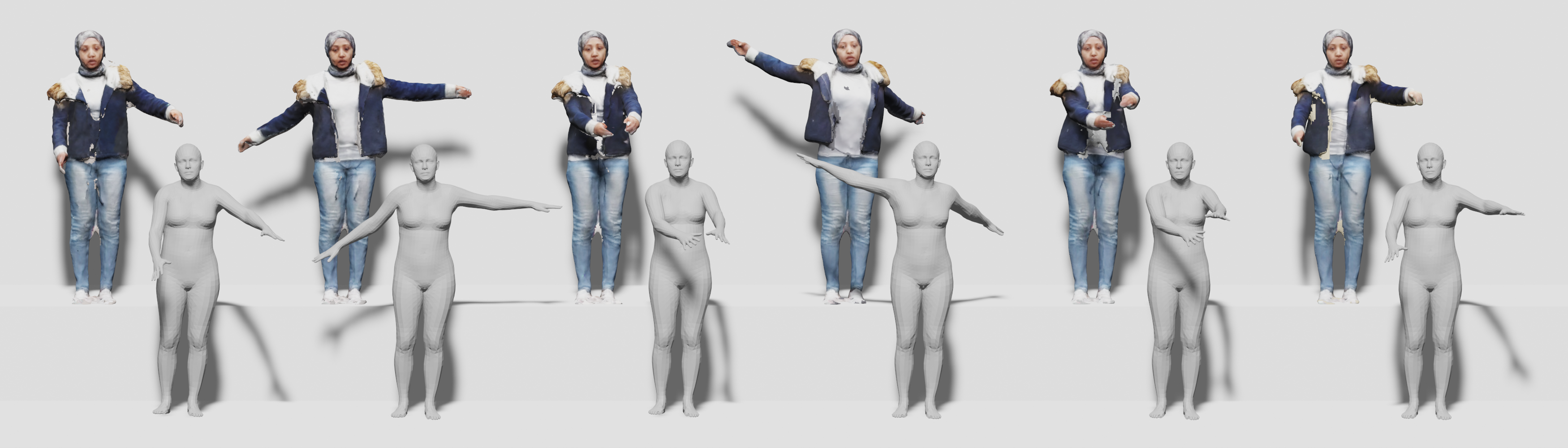}
    \caption{\name provides synchronized sequences of multiple data formats and annotations. Here we demonstrate textured mesh sequences and SMPL parameter sequences}
    \label{fig:textured_mesh_and_smpl}
\end{figure}

We select SMPL \cite{loper2015smpl} as the human parametric model for its popularity. There are two stages of registration (I and II). Stage I is used to obtain accurate shape parameters from the static high-resolution scan, whereas stage II is used to obtain pose parameters from the dynamic sequence, with shape parameters from stage I.
The registration is formulated as an optimization task to obtain SMPL pose parameters $\theta \in \mathbb{R}^{n \times 72}$, shape parameters $\beta \in \mathbb{R}^{n \times 10}$ (stage I only) and translation parameters $t \in \mathbb{R}^{n \times 3}$ where $n$ is the number of frames ($n=1$ for stage I), with the following energy terms and constraints. We show a sample sequence of SMPL models with reconstructed textured mesh in \Fig\ref{fig:textured_mesh_and_smpl}.

\noindent \textbf{Keypoint Energy.}
SMPLify \cite{bogo2016keep} estimates camera parameters to leverage 2D keypoint supervision, which may be prone to depth and scale ambiguity. Hence, we develop the keypoint energy on 3D keypoints. For simplicity, we denote $P_{3D}$ as $P$, the global rigid transformation derived from the SMPL kinematic tree as $\mathcal{T}$, the joint regressor as $\mathcal{J}$. We formulate the energy term:
%
\begin{equation}
    E_{\mathcal{P}}(\theta, \beta, t) = \frac{1}{|\mathcal{P}|}\sum_{i}^{|\mathcal{P}|}\|\mathcal{T}(\mathcal{J}(\beta)_{i}, {\theta}), t) - \mathcal{P}_{i}\|
\end{equation}

\begin{figure}[t]
    \centering
    \includegraphics[width=\linewidth]{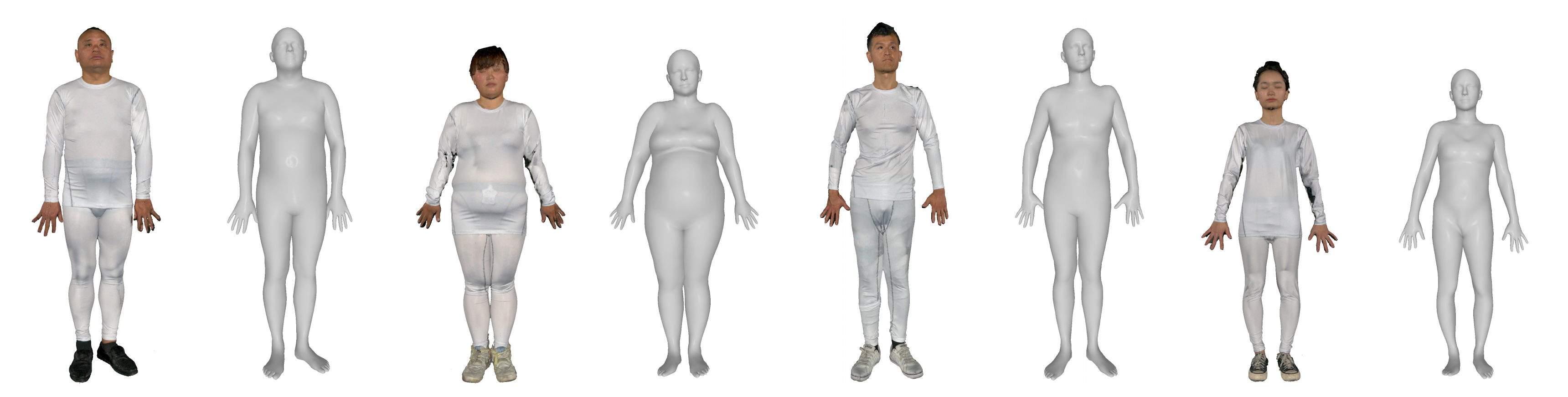}
    \caption{Examples of SMPL registered on high-resolution static body scans for accurate shape parameters. The subjects are instructed to wear tight clothes for this scan. Note that each subject has another naturally clothed scan}
    \label{fig:registration}
\end{figure}

\noindent \textbf{Surface Energy.}
To supplement 3D keypoints that do not provide sufficient constraint for shape parameters, we add an additional surface energy term for registration on the high-resolution minimally clothed scans in stage I only. We use bi-directional Chamfer distance to gauge the difference between two mesh surfaces:
%
\begin{equation}
    E_{S} = 
    \frac{1}{|\mathcal{V}_{H}|}\sum_{v_{H}\in\mathcal{V}_{H}}\underset{v_{S}\in\mathcal{V}_{S}}{\min}\|v_{H}-v_{S}\| + \frac{1}{|\mathcal{V}_{S}|}\sum_{v_{S}\in\mathcal{V}_{S}}\underset{v_{H}\in\mathcal{V}_{H}}{\min}\|v_{H}-v_{S}\|
    \label{eq:surface_energy}
\end{equation}
\noindent where $\mathcal{V}_{H}$ and $\mathcal{V}_{S}$ are the mesh vertices of the high-resolution scan and SMPL.

\noindent \textbf{Shape Consistency.}
Unlike existing work \cite{patel2021agora} that enforces an inter-beta energy term due to the lack of minimally clothed scan of each subject, we obtain accurate shape parameters from the high-resolution scan that allow us to apply constant beta parameters in the registration in stage II.

\noindent \textbf{Full-body Joint Angle Prior.}
Joint rotation limitations serve as an important constraint to prevent unnaturally twisted poses. We extend existing work \cite{bogo2016keep, pavlakos2019expressive} that only applies constraints on elbows and knees to all $J=23$ joints in SMPL. The constraint is formulated as a strong penalty outside the plausible rotation range (with more details included in the \Supp):
%
\begin{equation}
    E_a = \frac{1}{J\times3}\sum_{j}^{J\times3}exp(
    \max(\theta_{i} - \theta_{i}^{u}, 0) + 
    \max(\theta_{i}^{l} - \theta_{i}, 0)) - 1 
\end{equation}
\noindent where $\theta_{i}^{u}$ and $\theta_{i}^{l}$ are the upper and lower limit of a rotation angle. Note that each joint rotation is converted to three Euler angles which can be interpreted as a series of individual rotations to decouple the original axis-angle representation.

\subsection{Textured Mesh Reconstruction}
\label{sec:toolchain:textured_mesh_reconstruction}
\begin{figure}[t]
    \centering
    \includegraphics[width=\linewidth]{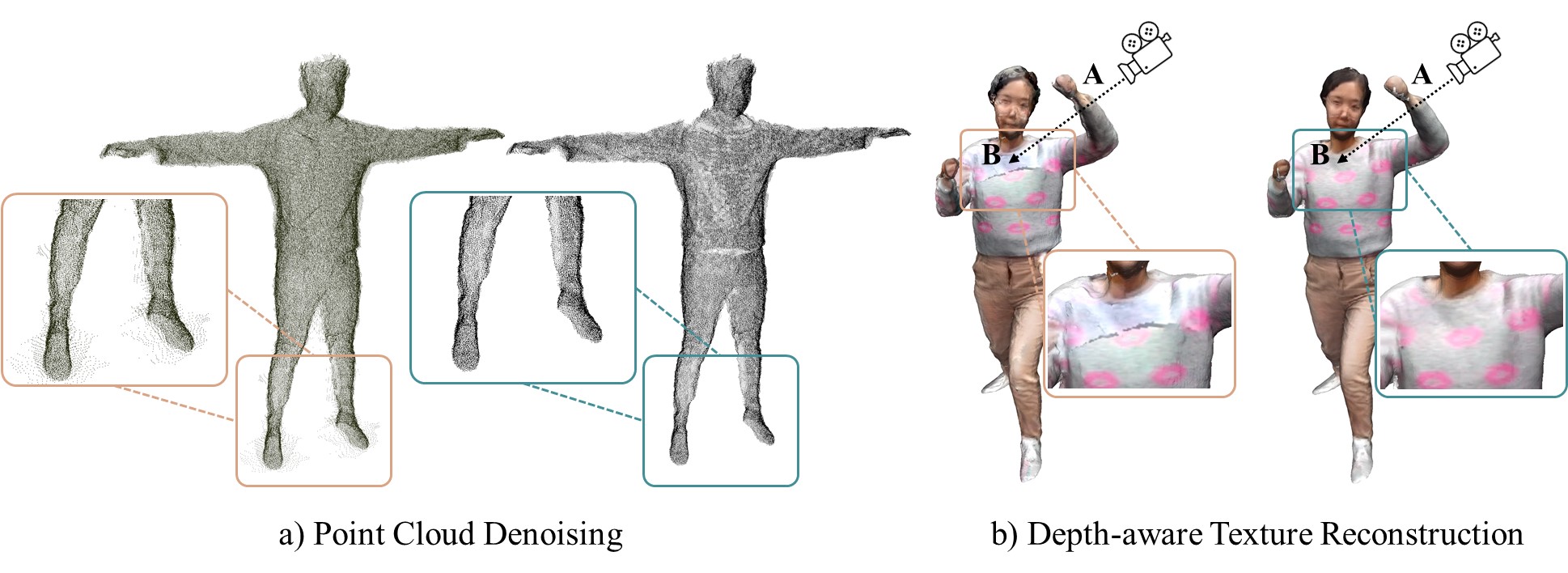}
    \caption{Key steps to textured mesh reconstruction. a) Point cloud denoising removes noisy points. b) Depth-aware texture reconstruction prevents texture miss projection artifacts (such as projecting texture at point A to point B) due to misalignment between the actual subject and the reconstructed geometry}
    \label{fig:textured_mesh_reconstruction}
\end{figure}

\noindent \textbf{Point Cloud Reconstruction and Denoising.}
We convert depth maps to point clouds and transform them into a world coordinate system with camera extrinsic parameters. However, the depth images captured by Kinect contain noisy pixels, which are prominent at subject boundaries where the depth gradient is large. To solve this issue, we first generate a binary boundary mask through edge finding with Laplacian of Gaussian Filters. Since our cameras have highly overlapped views to supplement points for one another, we apply a more aggressive threshold to remove boundary pixels. After the point cloud is reconstructed from the denoised depth images, we apply Statistical Outlier Removal ~\cite{hodge2004survey} to further remove sprinkle noises.

\noindent \textbf{Geometry and Depth-aware Texture Reconstruction.}
With complete and dense point cloud reconstructed, we apply Poisson Surface Reconstruction with envelope constraints~\cite{kazhdan2020poisson} to reconstruct the watertight mesh. However, due to inevitable self-occlusion in complicated poses, interpolation artifacts arise from missing depth information, which leads to a shrunk or a dilated geometry. These artifacts are negligible for geometry reconstruction. However, a prominent artifact appears when projecting a texture onto the mesh even if the inconsistency between the true surface and the reconstructed surface is small. Hence, we extend MVS-texturing ~\cite{Waechter2014Texturing} to be depth-aware in texture reconstruction. We render the reconstructed mesh back into the camera view and compare the rendered depth map with the original depth map to generate the difference mask. We then mask out all the misalignment regions where the depth difference exceeds a threshold $\tau_d$. The masked regions do not contribute to texture projection. As shown in \Fig\ref{fig:textured_mesh_reconstruction}(b), the depth-aware texture reconstruction is more accurate and visually pleasing. 

\section{Action Set}
\label{sec:action_set}
\begin{figure}[t]
  \includegraphics[width=\linewidth]{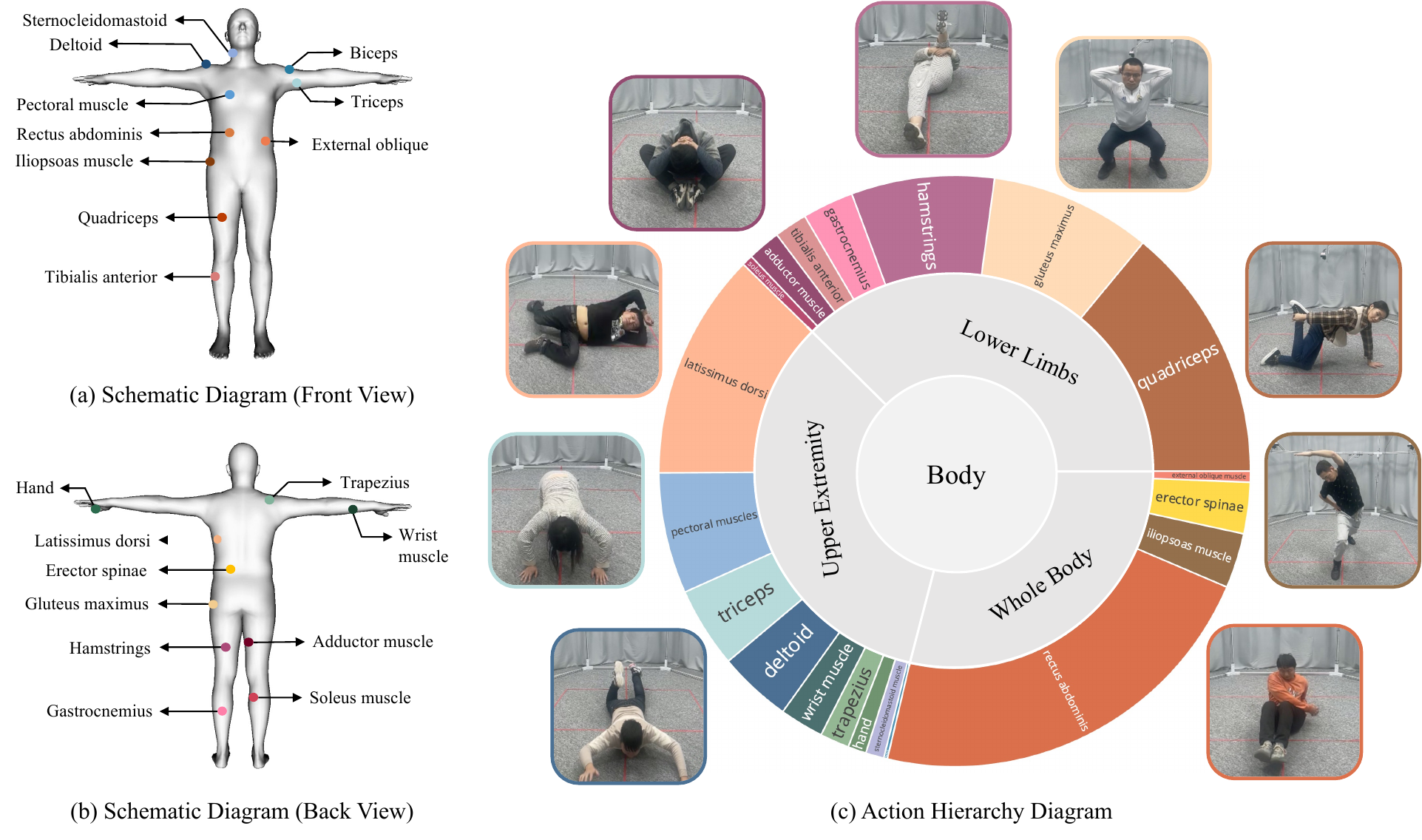}
  \caption{Schematic diagram of muscles from a) front and b) back views. c) \name categorizes 500 actions hierarchically, first by body parts to achieve \textit{complete} body coverage, then by driving muscles for \textit{unambiguous} action definition}
  \label{fig:stat_action}
\end{figure}


Understanding human actions is a long-standing computer vision task. In this section, we elaborate on the two principles, following which we design the action set of \numactions actions: \textit{completeness} and \textit{unambiguity}. More details are included in the \Supp.

\noindent \textbf{Completeness.}
We build the action set to cover plausible human movements as much as possible. Compared to the popular 3D action recognition dataset NTU-RGBD-120~\cite{liu2019ntu} whose actions are focused on upper body movements, we employ a hierarchical design to first divide possible actions into upper extremity, lower limbs, and whole-body movements. Such design allows us to achieve a balance between various body parts instead of over-emphasizing a specific group of movements. Note that we define whole body movements to be actions that require multiple body parts to collaborate, including different poses of the body trunk (\eg lying down and sprawling). \Fig\ref{fig:stat_action}(c) demonstrates the action hierarchy and examples of interesting actions that are vastly diverse.

\noindent \textbf{Unambiguity.}
Instead of providing a general description of the motions \cite{soomro2012ucf101,karpathy2014large,carreira2019short,monfort2019moments,ionescu2013human3,mehta2017monocular,von2018recovering}, we argue that the action classes should be clearly defined and are easy to identify and reproduce. Inspired by the fact that all human actions are the result of muscular contractions, we propose a \textit{muscle-driven} strategy to systematically design the action set from the perspective of human anatomy. As illustrated in \Fig\ref{fig:stat_action}(a)(b), $20$ major muscles are identified by professionals in fitness and yoga training, who then put together a list of standard movements associated with these muscles. Moreover, we cross-check with the action definitions from existing datasets~\cite{gu2018ava, carreira2019short, caba2015activitynet, chung2021haa500, karpathy2014large, li2022hake, liu2019ntu, joo2015panoptic} to ensure a wide coverage.

\section{Subjects}
\label{sec:subjects}
\begin{figure}[t]
    \centering
    \includegraphics[width=\linewidth]{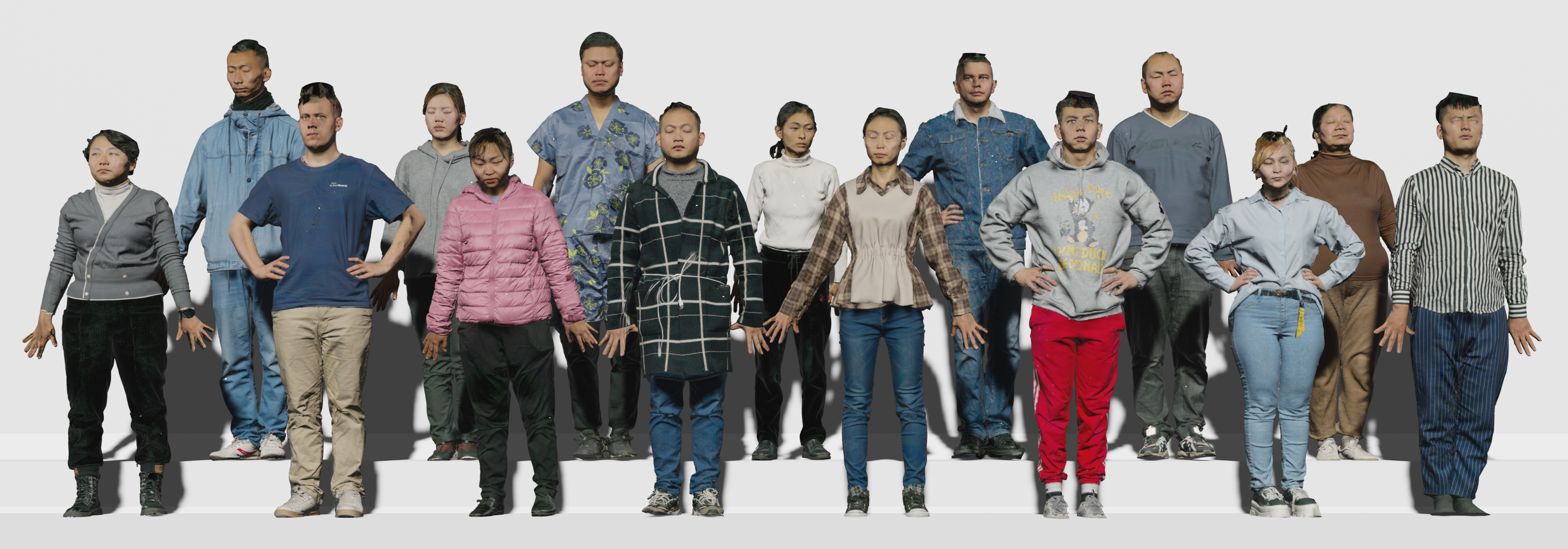}
    \caption{\name contains \numsubjects subjects with diverse appearances. For each subject, a naturally clothed high-resolution scan is obtained}
    \label{fig:subjects}
\end{figure}

\name consists of \numsubjects subjects with a wide coverage of genders, ages, body shapes (heights, weights), and ethnicity. The subjects are instructed to wear their personal daily clothes to achieve a large collection of natural appearances. We demonstrate examples of high-resolution scans of the subjects in \Fig\ref{fig:subjects}. We include statistics in the \Supp.







\section{Experiments}

In this section, we evaluate popular methods from various research fields on \name. To constrain the training within a reasonable computation budget, we sample 10\% of data and split them into training and testing sets for both Kinects and iPhone. The details are included in the \Supp.



\setlength{\tabcolsep}{4pt}

\begin{wraptable}{r}{0.5\textwidth}
\centering
\caption{\textbf{Action Recognition}}
\label{tab:action}
\begin{tabular}{lcc}
    \hline
    Method & Top-1 (\%)$\uparrow$ &Top-5 (\%)$\uparrow$ \\  
    \toprule
    ST-GCN & 72.5 & 94.3\\
    2s-AGCN & 74.1 & 95.4\\
    \bottomrule
\end{tabular}
\end{wraptable}

\setlength{\tabcolsep}{1.4pt}

\noindent \textbf{Action Recognition.}
\name provides action labels and 3D skeletal positions, which can verify its usefulness on 3D action recognition. 
%
Specifically, we train popular graph-based methods (STGCN \cite{yan2018spatial} and 2s-AGCN \cite{shi2019skeleton}) on \name. Results are shown in \Tab~\ref{tab:action}. Compared to NTU RGB+D, a large-scale 3D action recognition dataset and a standard benchmark that contains 120 actions~\cite{liu2019ntu}, \name may be more challenging since 2s-AGCN~\cite{shi2019skeleton} achieves Top-1 accuracy of 88.9\% and 82.9\% on NTU RGB+D 60 and 120 respectively, but 74.1\% only on \name. The difficulties come from the whole-body coverage design in our action set, instead of over-emphasis on certain body parts (\eg NTU RGB+D has a large proportion of upper body movements). Moreover, we observe a significant gap between Top-1 and Top-5 accuracy ($\sim$30\%). We attribute this phenomenon to the fact that there are plenty of \textit{intra-actions} in \name. For example, there are similar variants of push-ups such as quadruped push-ups, kneeling push-ups, and leg push-ups. This challenges the model to pay more attention to the fine-grained differences in these actions. Hence, we find \name would serve as an indicative benchmark for fine-grained action understanding. 



\setlength{\tabcolsep}{4pt}

\begin{wraptable}{r}{0.55\textwidth}
\centering
\caption{\textbf{3D Keypoint Detection}. PA: PA-MPJPE. Row 1-3: FCN~\cite{martinez2017simple}; Row 4-6: Video3D~\cite{pavllo20193d}}
\label{tab:3d_keypoint}
\begin{tabular}{ccccc}
\toprule
    Train & Test & MPJPE $\downarrow$ & PA $\downarrow$\\
    \midrule
     \name&\name&78.5&46.3\\
     H36M&AIST++&133.9&73.1\\
    \name&AIST++&116.4&67.2\\
    \midrule
    
     \name&\name&73.1&43.5\\
    H36M&AIST++&128.5&72.0\\
    \name&AIST++&109.2&63.5\\
    
    
  \bottomrule
\end{tabular}
\end{wraptable}

\setlength{\tabcolsep}{1.4pt}
\noindent \textbf{3D Keypoint Detection.}
With the well-annotated 3D keypoints, \name supports 3D keypoint detection. We employ popular 2D-to-3D lifting backbones~\cite{martinez2017simple,pavllo20193d} as single-frame and multi-frame baselines on \name.
%
%
We experiment with different training and test settings to obtain the baseline results in \Tab~\ref{tab:3d_keypoint}. First, in-domain training and testing on \name are provided. The values are slightly higher than the same baselines on Human3.6M~\cite{ionescu2013human3} (on which FCN obtains MPJPE of 53.4 mm). Second, methods trained on \name tend to generalize better than on Human3.6M. This may be attributed to \name's diverse collection of subjects and actions.

\setlength{\tabcolsep}{4pt}

\begin{wraptable}{r}{0.42\textwidth}
\centering
\caption{\textbf{3D Parametric Human Recovery}. Image- and point cloud-based methods are evaluated}
\label{tab:3d_human_recovery}
\begin{tabular}{lcc}
  \toprule
  Method & MPJPE $\downarrow$ & PA $\downarrow$ \\
  \midrule
  HMR & 54.78 & 36.14     \\
  VoteHMR & 144.99 & 106.32    \\
  \bottomrule
\end{tabular}
\end{wraptable}

\setlength{\tabcolsep}{1.4pt}

\noindent \textbf{3D Parametric Human Recovery.}
\name provides SMPL annotations, RGB and RGB-D sequences. Hence, we evaluate HMR \cite{kanazawa2018end}, not only one of the first deep learning approaches towards 3D parametric human recovery but a fundamental component for follow-up works \cite{kolotouros2019learning, kocabas2020vibe}, to represent image-based methods. In addition, we employ VoteHMR \cite{liu2021votehmr}, a recent work that takes point clouds as the input. In \Tab\ref{tab:3d_human_recovery}, we find that HMR has achieved low MPJPE and PA-MPJPE, which may be attributed to the clearly defined action set and the training set already includes all action classes. However, VoteHMR is not performing well. We argue that existing point cloud-based methods \cite{jiang2019skeleton, wang2021locally, liu2021votehmr} rely heavily on synthetic data for training and evaluation, whereas \name provides genuine point clouds from commercial RGB-D sensors that remain challenging.

\setlength{\tabcolsep}{4pt}

\begin{wraptable}{r}{0.5\textwidth}
  \centering
  \caption{\textbf{Geometry Reconstruction}}
  \label{tab:recon}
  \begin{tabular}{lccc}
    \toprule
    Method & PIFu & PIFuHD & F4D \\
    \midrule
    CD ($10^{-2}$ m) & 7.92 & 7.73 & 1.80 \\
    \bottomrule 
  \end{tabular}
\end{wraptable}

\noindent \textbf{Textured Mesh Reconstruction.}
We gauge mesh geometry reconstruction quality of PIFu, PIFuHD, and Function4D (F4D) in \Tab\ref{tab:recon} with Chamfer distance (CD) as the metric. 
Note that benefiting from the multi-modality signals, \name supports a wide range of surface reconstruction methods that leverage various input types like PIFu~\cite{saito2019pifu} (RGB-only), 3D Self-Portrait~\cite{Li2021Portrait} (single-view RGBD video), and CON~\cite{Peng2020ECCV_Full} (multi-view depth point cloud). 

\setlength{\tabcolsep}{4pt}

\begin{wraptable}{r}{0.56\textwidth}
\centering
\caption{
\textbf{Mobile Device}. The models are trained with different training sets, and evaluated on \name iPhone test set. Kin.: Kinect training set. iPh.: iPhone training set.
}
\label{tab:mobile_device}
\begin{tabular}{lcccc}
  \toprule
  Method & Kin. & iPh. & MPJPE $\downarrow$ & PA $\downarrow$\\
  \midrule
  HMR & \checkmark & -          & 97.81 & 52.74 \\
  HMR & - & \checkmark          & 72.62	& 41.86 \\
  VoteHMR & \checkmark & -      & 255.71 & 162.00 \\
  VoteHMR & - & \checkmark      & 83.18	& 61.69 \\
  \bottomrule
\end{tabular}
\end{wraptable}

\setlength{\tabcolsep}{1.4pt}

\noindent \textbf{Mobile Device.}
It is under-explored that if model trained with the regular device is readily transferable to the mobile device. In \Tab\ref{tab:mobile_device}, we study the performance gaps across devices. For the image-based method, we find that there exists a considerable domain gap across devices, despite that they have similar resolutions. Moreover, for the point cloud-based method, the domain gap is much more significant as the mobile device tends to have much sparser point clouds as a result of lower depth map resolution. Hence, it remains a challenging problem to transfer knowledge across devices, especially for point cloud-based methods.

\section{Discussion}
We present \name, a large-scale 4D human dataset that features multi-modal data and annotations, inclusion of mobile device, a comprehensive action set, and support for multiple tasks. Our experiments point out interesting directions that await future research, such as fine-grained action recognition, point cloud-based parametric human estimation, dynamic mesh sequence reconstruction, transferring knowledge across devices, and potentially, multi-task joint training. We hope \name would facilitate the development of better algorithms for sensing and modeling humans.

\noindent \textbf{Acknowledgements.} This work is supported by NTU NAP, MOE AcRF Tier 2 (T2EP20221-0012), NSFC No.62171255, and under the RIE2020 Industry Alignment Fund - Industry Collaboration Projects (IAF-ICP) Funding Initiative, as well as cash and in-kind contribution from the industry partner(s).

\clearpage

\bibliographystyle{splncs04}
\bibliography{references}

\clearpage
\appendix

\section{Overview}
We provide additional details of
data collection (\Sec\ref{sec:supp:data_collection}), 
hardware (\Sec\ref{sec:supp:hardware}), 
toolchain (\Sec\ref{sec:supp:toolchain}), 
action set (\Sec\ref{sec:supp:action_set}), 
subjects (\Sec\ref{sec:supp:subjects}),
experiments (\Sec\ref{sec:supp:experiments}), and
a more complete dataset comparison (\Sec\ref{sec:supp:dataset_comparison}).

\section{Additional Details of Data Collection}
\label{sec:supp:data_collection}

\begin{figure}[t]
    \centering
    \includegraphics[width=\linewidth]{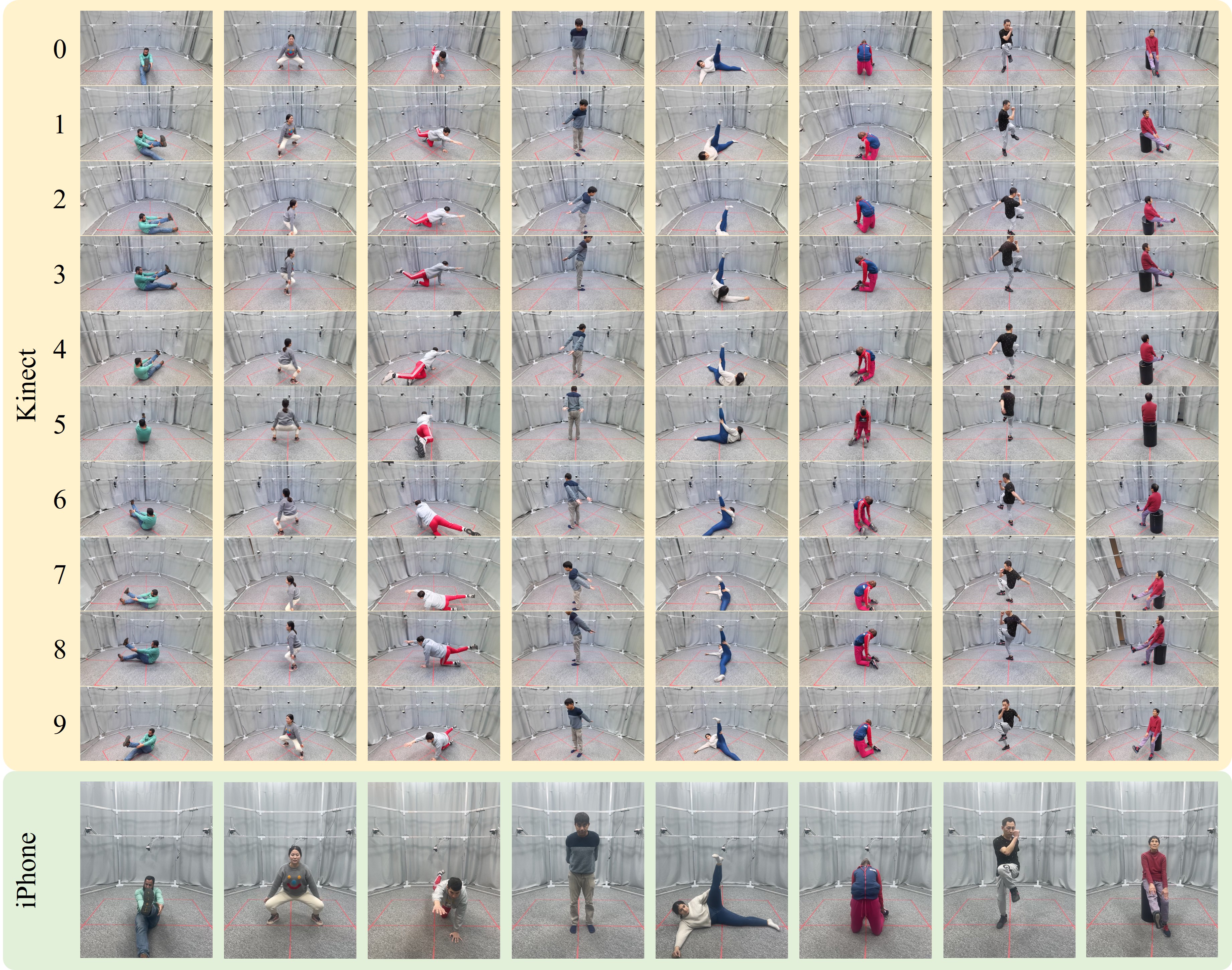}
    \caption{\name deploys ten Azure Kinects and an iPhone 12 Pro Max for multi-view sequential data collection. We show several synchronized RGB frames captured with our hardware setup. The numbers are device IDs}
    \label{fig:multiview_diversity}
\end{figure}

The data collection has two stages for each subject. \textbf{1)} each subject receives two high-resolution scans, one with natural clothes on and the other with a tight-fitting suit on, both captured by the Artex Eva 3D Scanner. To ensure the high quality of the scans, the subjects are instructed to stand in a special pose (the \textit{canonical pose}) on a turntable, that allows for a 360-degree full-body scanning with minimal self-occlusion. Each high-resolution scan includes an MTL information file, an OBJ mesh file, and a BMP texture file. \textbf{2)} After that static body scanning, the subject enters the framework and follows instructions to perform 40-60 actions, randomly sampled from the action set that contains 500 actions. Each action that a subject performs is a \textit{sequence}, that consists of ten Kinect RGB-D sequences and an iPhone RGB-D sequence. We show sample frames collected with our hardware setup in \Fig\ref{fig:multiview_diversity}. Each sequence takes 5-15 seconds and 150-450 frames at 30 FPS per view. We compress all sequential data in a custom data format \textit{SMC} that is developed based on HDF5 format. The SMC file also contains additional information such as camera parameters, subject ID, and action ID. 

\section{Additional Details of Hardware}
\label{sec:supp:hardware}

\subsection{Sensors}
We provide more details on the RGB-D sensor (Azure Kinect). We set operating mode to \textit{NFOV unbinned} for the depth cameras, which results in the largest view overlap with the color camera and the densest point clouds. The depth camera in this mode has an FOV of $90^{\circ} \times59^{\circ}$. The operating range of the depth sensor in this mode is between 0.5 m to 3.86 m. The typical systematic error of the depth sensor is less than 11 mm + 0.1\% of distance with a standard deviation of less than 17 mm. In view of the limited FOV and depth error-distance relation, we design our aluminum framework such that the subject is around 2 m away from the Kinects: at that distance, the FOV can accommodate the subject's whole body, without incurring any extra depth error.

\begin{figure}[t]
    \centering
    \includegraphics[width=0.9\linewidth]{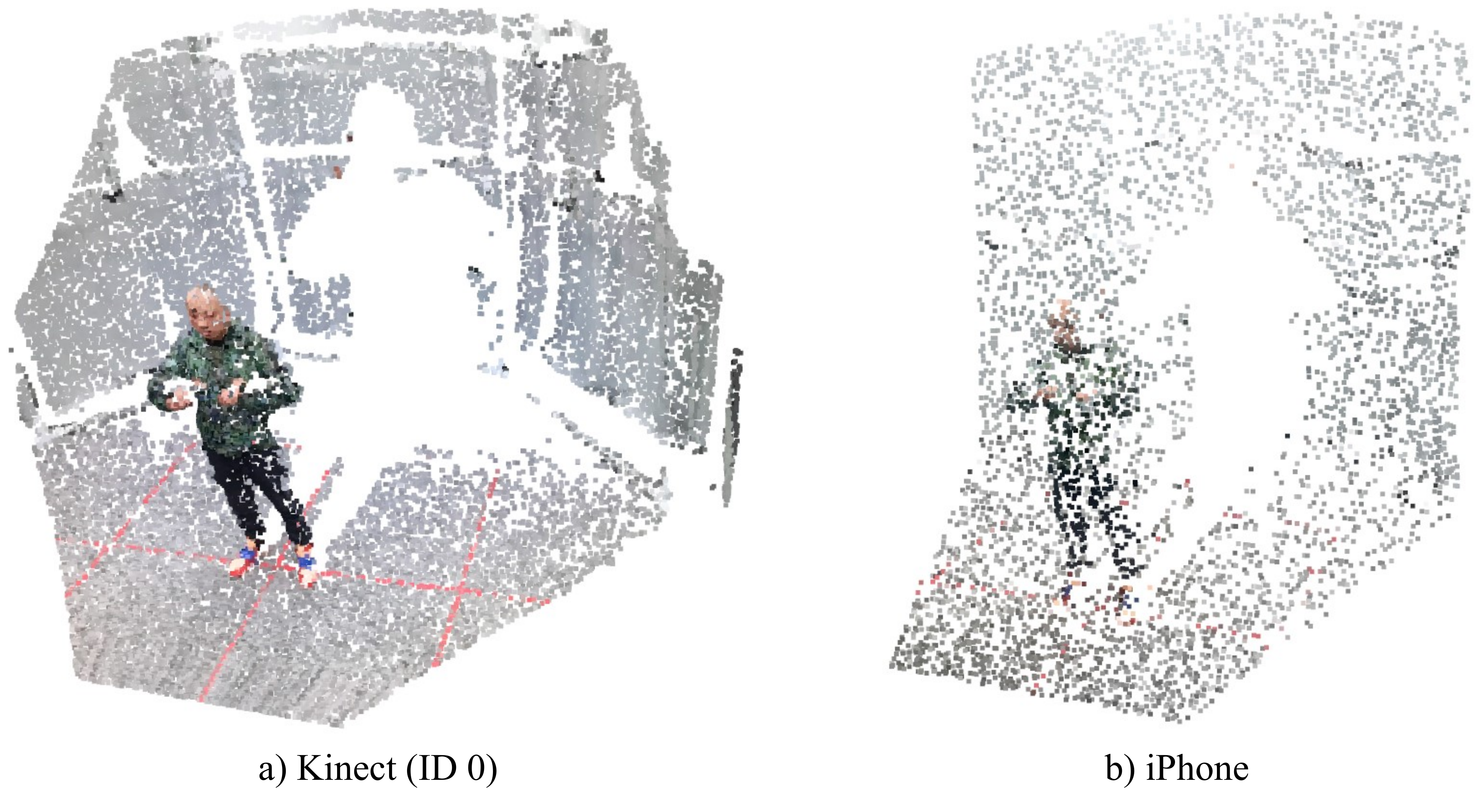}
    \caption{The point clouds produced by the Kinect and the iPhone are different: the latter is significantly sparser. Note that the point clouds shown here are raw (not filtered or denoised). For visual comparison purpose, both point clouds are downsampled by the same factor of 10}
    \label{fig:kinect_vs_iphone_pc}
\end{figure}

\subsection{Synchronization}
Our data sampling program runs on a workstation, and it 1) integrates the Kinect SDK, and 2) communicates with the iPhone app developed based on ARKit through TCP. Since there is no existing hardware approach to Kinect-iPhone synchronization, we develop a method to compute the difference between Kinect clock and iPhone ARKit clock $t_{K{\rightarrow}A}$.  Hence, we first obtain the offset from the workstation to the Kinects $t_{K{\rightarrow}W}$ as
$$t_{K{\rightarrow}W} = t_{W} - t_{K}$$ 
\noindent where $t_{K}$ is the Kinect clock time and $t_{W}$ is the workstation's system time, obtained at the same moment. We also send a message to the iPhone app, which records down the iPhone system clock $t_{I}$ upon receiving the message and sends back a message to the workstation to complete a round trip. We compute the offset from the iPhone system clock to the workstation system clock $t_{W{\rightarrow}I}$ as
$$t_{W{\rightarrow}I} = t_{I} - t_{W} - \frac{t_{round}}{2}$$ 
where $t_{round}$ is the round trip time taken. Note that there is an additional offset between the ARKit clock and the iPhone system clock $t_{I{\rightarrow}A}$, computed as 
$$t_{I{\rightarrow}A} = t_{A} - t_{I}$$
\noindent where $t_{A}$ is the ARKit clock. 
Finally, the required clock difference $t_{K{\rightarrow}A}$ is
$$t_{K{\rightarrow}A} = t_{K{\rightarrow}W} + t_{W{\rightarrow}I} + t_{I{\rightarrow}A}$$

\subsection{Point Clouds}
Both Kinect and iPhone produce depth maps that can be converted to point clouds. However, iPhone's point cloud is much sparser than Kinect's. We show unprocessed raw point clouds produced by the two types of sensors in \Fig\ref{fig:kinect_vs_iphone_pc}. In addition, iPhone does not report the LiDAR accuracy; we empirically find that iPhone point clouds are noisier, especially at the object boundaries, than Kinect point clouds.

\section{Additional Details of Toolchain}
\label{sec:supp:toolchain}

\subsection{Keypoint Annotation}
The overall pipeline for keypoint annotation is summarized in Algorithm \ref{alg:kp_annot}. 
\begin{algorithm}[h]
 \caption{Keypoint Annotation}
 \begin{algorithmic}[1]
 \renewcommand{\algorithmicrequire}{\textbf{Input:}}
 \renewcommand{\algorithmicensure}{\textbf{Output:}}
 \Require Detected 2D Keypoints $\hat{\cP}_{2D}$, camera parameters set $\cC$, keypoint threshold $\tau_k$, reprojection minimal threshold $\tau_{min}$, reprojection maximum threshold $\tau_{max}$, camera threshold step $\Delta_c$, best camera number $N_c$.
 \Ensure 3D Keypoints $\cP_{3D}$, 2D Keypoints $\cP_{2D}$

   \State $\tau_c = \tau_{min}$, $\hat{\cC} = \emptyset$
   \State $\bar{\cP}_{2D}$ = \Call{FilterKeypoints}{$\hat{\cP}_{2D}$, $\tau_k$}
   \While{$\tau_c \leq \tau_{max}$}
        \State $\cP_{3D}$ = \Call{Triangulate}{$\bar{\cP}_{2D}$, $\cC$}
        \State $\cP_{2D}$ = \Call{Reprojection}{$\cP_{3D}$}
        \While{$\tau_c \leq \tau_{max}$ and $|\hat{\cC}| < 3$}
            \State $\hat{\cC}$ = \Call{SelectCam}{$\cP_{2D}$, $\bar{\cP}_{2D}$, $\tau_c$, $N_c$}
            \State $\tau_c = \tau_c + \Delta_c$
        \EndWhile
        \If{$\cC$ == $\hat{\cC}$}
            \State \Return $\cP_{3D}$, $\cP_{2D}$
        \Else
            \State $\cC$ = $\hat{\cC}$
        \EndIf
   \EndWhile
   \State \Return Fail
 \end{algorithmic}
 \label{alg:kp_annot}
\end{algorithm}

\subsection{Full-body Angle Prior}
It is surprisingly difficult to find literature that provides a complete analysis of joint movement ranges, especially rotations in three degrees of freedom (DOF). Hence, we take references from artists' guidelines on human anatomy\footnote{\url{https://design.tutsplus.com/articles/human-anatomy-fundamentals-flexibility-and-joint-limitations--vector-25401}} and 3D modelers' suggested practices\footnote{\url{https://wiki.secondlife.com/wiki/Suggested_BVH_Joint_Rotation_Limits}}, to simplify the constraint such that the three DOF movement range is bounded by the maximum ranges in each of the DOF. Despite that this formulation is not perfect, it provides constraints that are otherwise completely absent. To easily apply the per-axis ranges, we convert the axis-angle representation into Euler angles and define the Z-axis to be aligned with the child bone of the joint in the kinematic tree (for example, \textit{forearm} is the child bone of the joint \textit{elbow}). To circumvent \textit{gimbal lock} as much as possible, we define the joint frame coordinate such that the second rotation axis (Y-axis) always falls on the less flexible axis (for which the rotation is unlikely to reach 90$^{\circ}$). Hence, we define the X-axis as the axis around which the largest rotation is achieved. Y-axis is finally defined with X- and Z-axis fixed. All values undergo manual inspection and are adjusted empirically. Note that the Euler angle rotation is used to generate a loss only; the joint rotation is still in axis-angle representation.

\subsection{Annotation Quality of SMPL Parameters.}
\begin{figure}[t]
  \centering
  \includegraphics[width=0.3\linewidth]{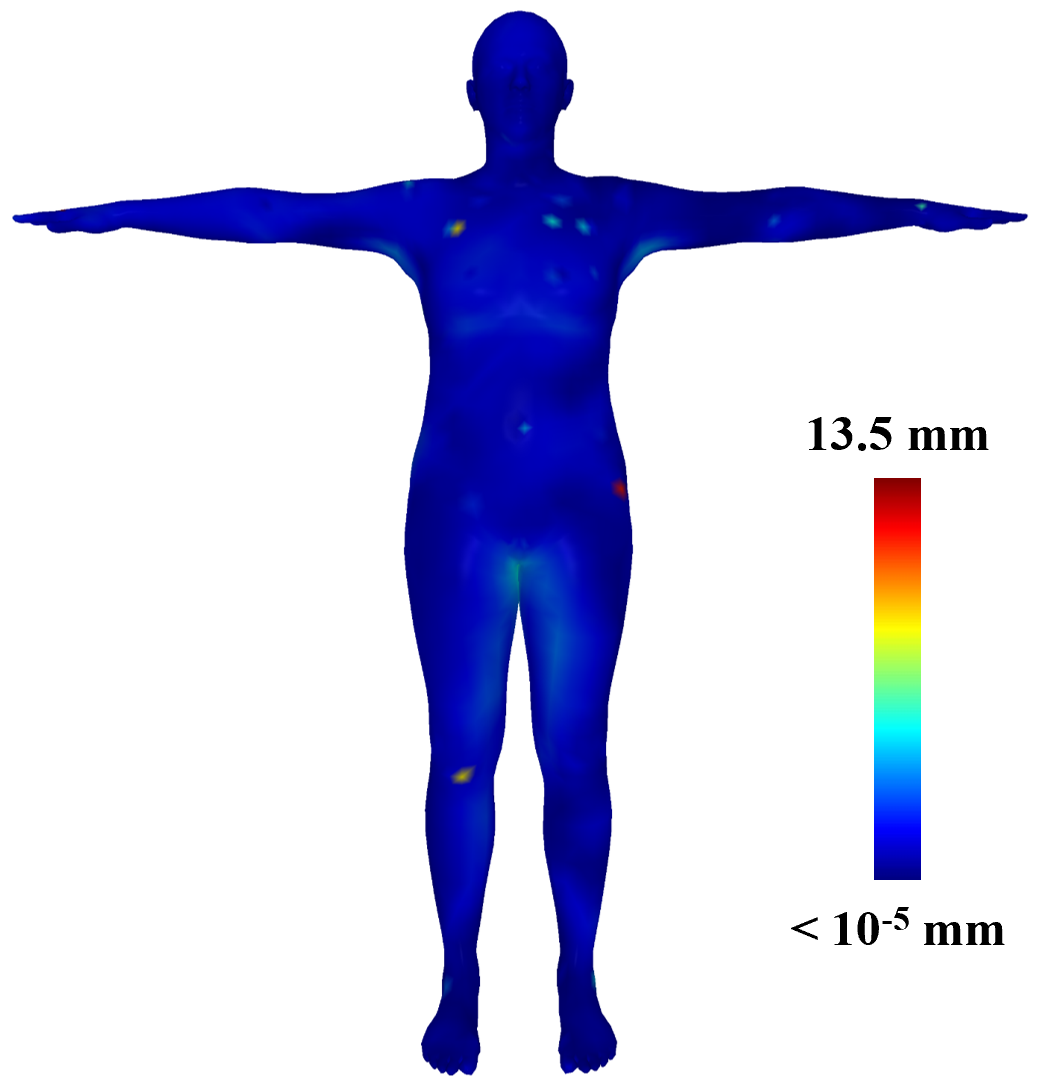}
  \caption{The registration accuracy on high-resolution mesh (minimally clothed). The metric is mean uni-directional Chamfer distance (from SMPL vertices to high-resolution mesh vertices). Our registration (and subsequently the body shape obtained) is mostly accurate}
  \label{fig:hr_mesh_registration_accuracy}
\end{figure}

To evaluate the body shape, we compute the per-vertex error on the high-resolution scan that is the uni-directional Chamfer distance from registered SMPL mesh vertices to the high-resolution scan vertices. Note that high-resolution scans have been scaled to the real height of scanned persons. The mean per-vertex error is 0.16 mm. We also visualize the registration quality in \Fig\ref{fig:hr_mesh_registration_accuracy}. To evaluate the body pose, we compute the per-joint error as the L2 Euclidean distance between 3D keypoints and 3D joints of registered SMPL on the dynamic sequences. The mean per-joint error is 38.18 mm. Note that the error is largely attributed to the difference in the joint definition of the keypoint detector and the parametric model. As a reference, registration with an accurate optical marker system \cite{ionescu2013human3, loper2014mosh} yields a per-joint error of 29.34 mm. 

\clearpage
\section{Additional Details of Action Set}
\label{sec:supp:action_set}
\begin{figure}[t]
  \includegraphics[width=\linewidth]{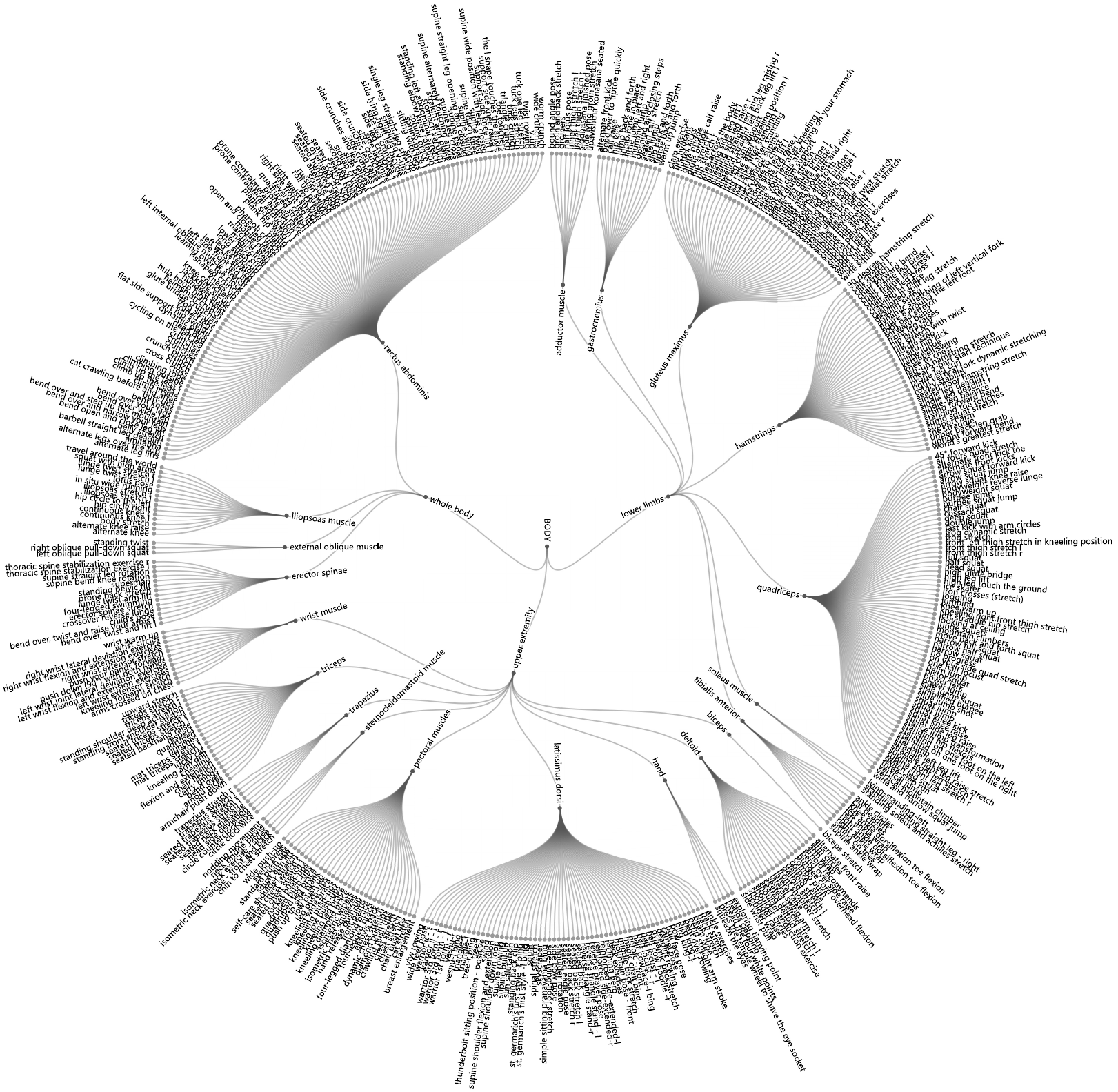}
  \caption{The complete set of \numactions actions}
  \label{fig:action_set_complete}
\end{figure}

\vspace{2mm}
\noindent \textbf{Design Process.} In \name, we design a hierarchical structure for a systematic coverage of different body parts to collate a \textit{complete} and \textit{unambiguous} action set. Specifically, we have \textit{body} at the center as the first order. The second order consists of \textit{whole body}, \textit{upper extremity} and \textit{lower limbs} that categorize actions by major body parts. After that, we propose a \textit{muscle-driven} strategy to further split each major body part into main muscle groups according to human anatomy as the third order. Finally, we involve domain experts to design a series of action variants associated with each muscle in the fourth order. The full action hierarchy is demonstrated in \Fig\ref{fig:action_set_complete}.

\vspace{2mm}
\noindent \textbf{Motion Diversity.} As \name contains a large amount of data, we further conduct a preliminary study on the motion diversity for further research on the motion prior learning. Specifically, We compute the mean standard deviation of joint angles of three datasets: 3DPW (0.159), AMASS (0.208), and \name (\textbf{0.269}). The higher mean standard deviation indicate higher diversity in motions. Although the AMASS dataset with a large-scale MoCap data is wildly used in many recent works to pretrain models, \name has more diversity in joint angles, showing its potential for human motion-related tasks.

\section{Additional Details of Subjects}
\label{sec:supp:subjects}

\begin{figure}[t]
    \centering
    \includegraphics[width=\linewidth]{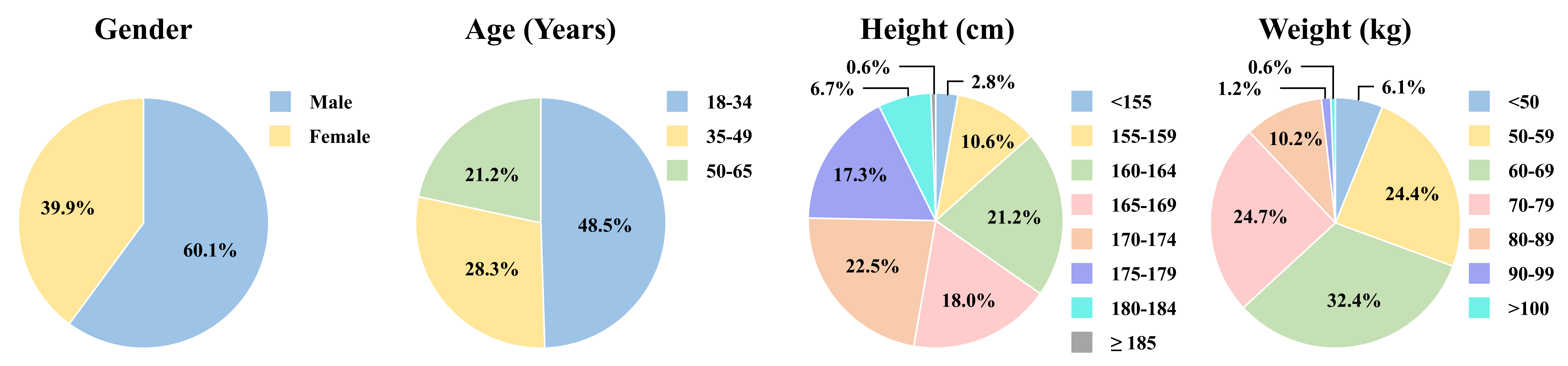}
    \caption{Statistics of \name subjects}
    \label{fig:subject_stats}
\end{figure}

\vspace{2mm}
\noindent \textbf{Statistics.}
\name  consists of \numsubjects subjects. To evaluate the diversity, we include key statistics (gender, age, height and weight) of the subjects in \Fig\ref{fig:subject_stats}.

\vspace{2mm}
\noindent \textbf{Ethics.}
\name involves a large number of human subjects so that we pay special attention to address ethic concerns. The recruitment process is conducted on an entirely voluntary basis. Actors and actresses who participate in \name are well-informed, with legal agreements signed to acknowledge that the data will be made public for research purposes. 

\section{Additional Details of Experiments}
\label{sec:supp:experiments}

\subsection{Splits and Protocols}
\name contains a massive scale of subjects (\numsubjects), actions (\numactions), sequences (\numsequences) and frames (\numframes). To constrain training and testing within a reasonable computation budget, we sample only 10\% of the data. We then develop three protocols to split iPhone and Kinect data into training and test sets. \textbf{Protocol 1 (P1)}: split by subjects, the training and test set are mutually exclusive and contain 70\% and 30\% of the subjects respectively. P1 is used for all experiments in the main paper. \textbf{Protocol 2 (P2)}: split by actions. We split actions into three categories according to major body parts involved: \textit{upper extremity}, \textit{lower limbs}, and \textit{whole body}. Training is conducted on one category whereas the test is conducted on the other two. \textbf{Protocol 3 (P3)}: split by views. Model is trained on only one view (the \textit{front} view, or the view of the iPhone and the Kinect with ID 0) and tested on all views.

\subsection{2D Keypoint Detection}

\setlength{\tabcolsep}{4pt}

\begin{wraptable}{r}{0.45\textwidth}
\vspace{-8.8mm}
\centering
\caption{2D Keypoint Detection under Protocol 1. Input image is resized to 384$\times$288}
\vspace{-2mm}
\label{tab:2d_keypoint}
\begin{tabular}{lcc}
  \toprule
  Method & AP$^{50} \uparrow$ & AP$^{75}\uparrow$ \\
  \midrule
  CPN~\cite{chen2018cascaded}  & 0.86 & 0.93 \\
  HRNet~\cite{sun2019hrnet}    & 0.91 & 0.97 \\
  Lite-HRNet~\cite{yu2021lite} & 0.87 & 0.93 \\
  \bottomrule
\end{tabular}
\vspace{-5mm}
\end{wraptable}

\setlength{\tabcolsep}{1.4pt}

We study 2D keypoint detection baselines on \name primarily for 2D-to-3D keypoint lifting. CPN \cite{chen2018cascaded} is a cascaded pyramid network to improve hard keypoints detection. HRNet \cite{sun2019hrnet} is a novel high-resolution network that obtains high performance on COCO dataset~\cite{lin2014microsoft}, and LiteHRNet is an efficient version of HRNet. The comparison results are listed in \Tab~\ref{tab:2d_keypoint}. Because 2D keypoints are often used as an intermediate representation of 3D keypoints in a two-stage manner~\cite{martinez2017simple,pavllo20193d}, the good performance in this task can be helpful to the estimation of subsequent 3D.

\subsection{3D Keypoint Detection}
\setlength{\tabcolsep}{4pt}

\begin{table}[t]
\centering
\caption{3D keypoint detection under Protocol 2 on Kinect splits. FCN is used as the base model.}
\label{tab:3d_keypoint_p2}
\vspace{2mm}
\begin{tabular}{llcc}
  \toprule
  Training & Testing & MPJPE $\downarrow$ & PA-MPJPE $\downarrow$ \\
  \midrule
  Lower Limbs & Upper Extremity  & 70.3 & 55.7  \\
  Lower Limbs & Whole Body       & 97.5 & 72.3  \\
  Upper Extremity & Lower Limbs  & 75.8 & 55.1 \\
  Upper Extremity & Whole Body  & 99.6 & 72.5  \\
  Whole Body & Lower Limbs       & 77.4 & 56.2 \\
  Whole Body & Upper Extremity  & 86.2 & 66.4 \\
  \midrule
  \multicolumn{2}{c}{Mean Error} & 84.4 & 63.0\\
  \bottomrule
  
\end{tabular}
\end{table}

\setlength{\tabcolsep}{1.4pt}

\setlength{\tabcolsep}{4pt}

\begin{table}[t]
\centering
\caption{3D keypoint detection under Protocol 3 on Kinect splits. FCN is used as the base model. The model is trained on View 0 and tested on all views.}
\label{tab:3d_keypoint_p3}
\vspace{2mm}
\resizebox{\textwidth}{!}{
\begin{tabular}{lccccccccccc}
  \toprule
  View & 0 & 1 & 2 & 3 & 4 
  & 5 & 6 & 7 & 8 & 9 & Mean \\
  \midrule
  MPJPE $\downarrow$ & 66.4 & 97.2 & 167.1 & 172.0 & 247.2 
  & 268.4 & 245.1 &  175.3 & 165.4 & 95.9 & 170.0	 \\
  
  PA-MPJPE $\downarrow$ & 41.2 & 67.5 & 100.9 & 103.5 & 112.3
  & 118.7 & 111.8 &  103.9  & 100.2 & 67.1 & 92.7 \\
  \bottomrule
\end{tabular}}
\end{table}

\setlength{\tabcolsep}{1.4pt}

3D keypoint detection benchmarks under P1 setting are presented in the main paper and additional benchmarks under P2 and P3 are provided here. In \Tab\ref{tab:3d_keypoint_p2}, we show results on the cross-action (P2) performance of the FCN method~\cite{martinez2017simple}. Compared with Protocol 1, we observe that training with fewer actions and testing on unseen actions degrade the precision significantly, especially for cross-evaluation on the \textit{whole body} category which seems to have a large action distribution misalignment with the other two categories. Furthermore, we report results of cross-view (P3) in \Tab\ref{tab:3d_keypoint_p3}. When the model is only trained on one view (\ie, View 0), we observe a considerable domain gap across different views as the errors increase as the deviation from the test view from the training view increases. The experiment results indicate that cross-view 3D keypoint detection is challenging.

\subsection{3D Parametric Human Recovery}
\setlength{\tabcolsep}{4pt}

\begin{table}[t]
\centering
\caption{3D parametric human recovery under Protocol 2 on Kinect splits. HMR is used as the base model.}
\label{tab:3d_human_recovery_p2}
\vspace{2mm}
\begin{tabular}{llcc}
  \toprule
  Training & Testing & MPJPE $\downarrow$ & PA-MPJPE $\downarrow$ \\
  \midrule
  Lower Limbs & Upper Extremity  & 77.2 & 57.0  \\
  Lower Limbs & Whole Body       & 109.8 & 77.9  \\
  Upper Extremity & Lower Limbs  & 80.6 & 56.5 \\
  Upper Extremity & Whole Body  & 114.2 & 73.3  \\
  Whole Body & Lower Limbs       & 85.4 & 61.9 \\
  Whole Body & Upper Extremity  & 98.3 & 72.6 \\
  \midrule
  \multicolumn{2}{c}{Mean Error} & 94.2 & 66.5 \\
  \bottomrule
  
\end{tabular}
\end{table}

\setlength{\tabcolsep}{1.4pt}
\setlength{\tabcolsep}{4pt}

\begin{table}[t]
\centering
\caption{3D parametric human recovery under Protocol 3 on Kinect splits. HMR is used as the base model. The model is trained on View 0 and tested on all views.}
\label{tab:3d_human_recovery_p3}
\vspace{2mm}
\resizebox{\textwidth}{!}{
\begin{tabular}{lccccccccccc}
  \toprule
  View & 0 & 1 & 2 & 3 & 4 
  & 5 & 6 & 7 & 8 & 9 & Mean \\
  \midrule
  MPJPE $\downarrow$ &61.9 & 122.9 & 223.9 & 206.2 & 343.9 
  & 421.0 & 334.0 &  208.0  & 199.0 &  123.5 & 224.4 \\
  
  PA-MPJPE $\downarrow$ & 40.2 & 71.9 & 123.7 & 115.0 & 124.4 
  & 133.1 & 127.2 &  123.1  & 118.0 & 73.3 & 105.0 \\
  \bottomrule
\end{tabular}}
\end{table}

\setlength{\tabcolsep}{1.4pt}




In addition to P1 benchmarks for 3D parametric human recovery presented in the main paper, we also provide more benchmarks under P2 and P3. In \Tab\ref{tab:3d_human_recovery_p2}, we evaluate the cross-action (P2) performance of the HMR baseline. We find that testing on unseen poses is challenging (compared to P1 benchmark results). Moreover, \textit{whole body} actions seem to have a distribution that is further away from \textit{lower limbs} and \textit{upper extremity} actions. In \Tab\ref{tab:3d_human_recovery_p3}, we study the cross-view setting (P3), which is even worse than the cross-action setting. The HMR baseline is trained on View 0, and gives a clear trend that the greater the viewing angle difference, the larger the errors. View 5 is directly opposite View 0 and yields the largest error.

\clearpage
\subsection{Textured Mesh Reconstruction}
\begin{figure}[t]
  \includegraphics[width=\linewidth]{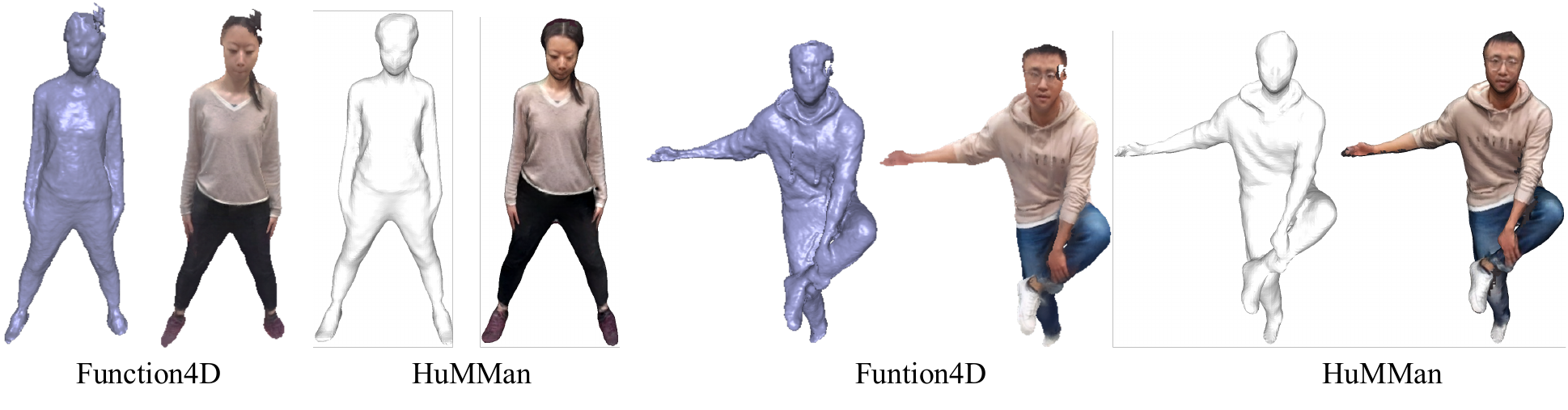}
  \caption{We compare Function4D with \name in textured mesh reconstruction}
  \label{fig:function_4d_baseline}
\end{figure}

To fully demonstrate the capacity of \name, we also provide the results of Function4D~\cite{tao2021function4d} as a baseline for textured mesh reconstruction since it combines both volumetric fusion and implicit surface reconstruction for volumetric capture in real-time. The results of Function4D, using 4 (ID: 0,3,6,9) views, are shown in \Fig\ref{fig:function_4d_baseline}.

\section{A More Complete Dataset Comparison}
\label{sec:supp:dataset_comparison}
In \Tab\ref{tab:dataset_comparison_complete}, we provide a more thorough comparison of \name with similar datasets for 1) action recognition, 2) 2D and 3D keypoint detection, 3) 3D parametric human recovery, and 4) mesh reconstruction.
\begin{table}[t]
\caption{A more complete comparison of \name with published datasets. Subj: subjects; Act: actions; Seq: sequences; Video: sequential data, not limited to RGB sequences; Mobile: mobile device in the sensor suite; D/PC: depth image or point cloud, only genuine point cloud collected from depth sensors are considered; Act: action label; K2D: 2D keypoints; K3D: 3D keypoints; Param: statistical model (\eg SMPL) parameters; Txtr: texture. -: not applicable or not reported}
\label{tab:dataset_comparison_complete}
\begin{center}
\scriptsize
\resizebox{\textwidth}{!}{
\begin{tabular}{lcccccccccccccc}

  \toprule

  \multirow{2}{*}{Dataset} & 
  \multirow{2}{*}{\#Subj} & 
  \multirow{2}{*}{\#Act} & 
  \multirow{2}{*}{\#Seq} & 
  \multirow{2}{*}{\#Frame} & 
  \multirow{2}{*}{Video} & 
  \multirow{2}{*}{Mobile} & 
  \multicolumn{8}{c}{Modalities} \\
  \cmidrule{8-15}
  &&&&&&& RGB & D/PC & Act & K2D & K3D & Param & Mesh & Txtr \\
  
  
  \midrule
  \multicolumn{15}{c}{Action Recognition} \\
  \midrule
  
  HMDB51~\cite{kuehne2011hmdb} &
  - & 51 & 7k & - & \checkmark & - &
  \checkmark & - & \checkmark & - & - & - & - & - \\
  
  UCF101~\cite{soomro2012ucf101} & 
  - & 101 & 13k & - & \checkmark & - & 
  \checkmark & - & \checkmark & - & - & - & - & - \\
  
  Sports1M~\cite{karpathy2014large} & 
  - & 487 & 1M & - & \checkmark & - & 
  \checkmark & - & \checkmark & - & - & - & - & - \\

  AVA~\cite{gu2018ava} & 
  - & 80 & 437 & - & \checkmark & - & 
  \checkmark & - & \checkmark & - & - & - & - & - \\
  
  Kinectics 700~\cite{carreira2019short} & 
  - & 700 & 650k & - & \checkmark & - & 
  \checkmark & - & \checkmark & - & - & - & - & - \\

  HACS~\cite{zhao2019hacs} & 
  - & 200 & 1.55M & - & \checkmark & - & 
  \checkmark & - & \checkmark & - & - & - & - & - \\

  Moments-In-Time~\cite{monfort2019moments} & 
  - & 339 & 1M & - & \checkmark & - & 
  \checkmark & - & \checkmark & - & - & - & - & - \\
  
  FineGym~\cite{shao2020finegym} & 
  - & 530 & 32k & - & \checkmark & - & 
  \checkmark & - & \checkmark & - & - & - & - & - \\

  HAA500~\cite{chung2021haa500} & 
  - & 500 & 10k & 591k & \checkmark & - & 
  \checkmark & - & \checkmark & - & - & - & - & - \\

  \midrule
  
  MSR-Action3D~\cite{li2010action} & 
  10 & 20 & 567 & - & \checkmark & - & 
  - & \checkmark & \checkmark & - & \checkmark & - & - & - \\
  
  Northwestern-UCLA~\cite{wang2014cross} & 
  10 & 10 & 1.47k & $>$23k & \checkmark & - & 
  \checkmark & \checkmark & \checkmark & - & \checkmark & - & - & - \\
  
  SYSU 3DHOI~\cite{hu2015jointly} & 
  40 & 12 & 65 & - & \checkmark & - & 
  \checkmark & \checkmark & \checkmark & - & \checkmark & - & - & - \\
  
  NTU RGB+D~\cite{shahroudy2016ntu} & 
  40 & 60 & 56k & - & \checkmark & - & 
  \checkmark & \checkmark & \checkmark & - & \checkmark & - & - & - \\
  
  NTU RGB+D 120~\cite{liu2019ntu} & 
  106 & 120 & 114k & - & \checkmark & - & 
  \checkmark & \checkmark & \checkmark & - & \checkmark & - & - & - \\
  
  NTU RGB+D X~\cite{trivedi2021ntu} & 
  106 & 120 & 113k & - & \checkmark & - & 
  \checkmark & \checkmark & \checkmark & - & \checkmark & \checkmark & - & - \\

  \midrule
  \multicolumn{15}{c}{2D/3D Keypoint Detection and 3D Parametric Human Recovery} \\
  \midrule

  J-HMDB~\cite{jhuang2013towards} & 
  - & 21 & 928 & 33.18k & \checkmark & - & 
  \checkmark & - & \checkmark & \checkmark & - & - & - & - \\
    
  Penn Action~\cite{zhang2013actemes} & 
  - & 15 & 2.32k & - & \checkmark & - & 
  \checkmark & - & \checkmark & \checkmark & - & - & - & - \\
  
  MPII~\cite{andriluka20142d} & 
  - & 410 & - & 24k & - & - & 
  \checkmark & - & \checkmark & \checkmark & - & - & - & - \\
  
  COCO~\cite{lin2014microsoft} & 
  - & - & - & 104k & - & - & 
  \checkmark & - & - & \checkmark & - & - & - & - \\
  
  PoseTrack~\cite{andriluka2018posetrack} & 
  - & - & $>$1.35k & $>$46k & \checkmark & - & 
  \checkmark & - & - & \checkmark & - & - & - & - \\  

  \midrule
  
  Human3.6M~\cite{ionescu2013human3} & 
  11 & 17 & 839 & 3.6M & \checkmark & - & 
  \checkmark & \checkmark & \checkmark & \checkmark & \checkmark & - & - & - \\ 
  
  CMU Panoptic~\cite{joo2015panoptic} & 
  8 & 5 & 65 & 154M & \checkmark & - & 
  \checkmark & \checkmark & - & \checkmark & \checkmark & - & - & - \\
  
  MPI-INF-3DHP~\cite{mehta2017monocular} & 
  8 & 8 & 16 & 1.3M & \checkmark & - & 
  \checkmark & - & - & \checkmark & \checkmark & - & - & - \\
  
  TotalCapture~\cite{trumble2017total} & 
  5 & 5 & 60 & 1.89M & \checkmark & - & 
  \checkmark & - & - & \checkmark & \checkmark & - & - & - \\ 
      
  3DPW~\cite{von2018recovering} & 
  7 & - & 60 & 51k & \checkmark & \checkmark & 
  \checkmark & - & - & - & - & \checkmark & - & - \\ 
      
  AMASS~\cite{mahmood2019amass} & 
  344 & - & $>$11k & $>$16.88M & \checkmark & - & 
  - & - & - & - & \checkmark & \checkmark & - & - \\ 

  Mirrored-Human~\cite{fang2021mirrored} & 
  - & 56 & 56 & $>$1.5M & \checkmark & - & 
  - & - & \checkmark & \checkmark & \checkmark & \checkmark & - & - \\

  AIST++~\cite{li2021ai} & 
  30 & - & 1.40k & 10.1M & \checkmark & - & 
  \checkmark & - & - & \checkmark & \checkmark & \checkmark & - & - \\

  \midrule
  \multicolumn{15}{c}{Mesh Reconstruction} \\
  \midrule

  ZJU LightStage~\cite{peng2021neural} & 
  6 & 6 & 9 & $>$1k & \checkmark & - & 
  \checkmark & - & \checkmark & \checkmark & \checkmark & \checkmark & \checkmark & \checkmark \\

  CAPE~\cite{ma2020learning} & 
  15 & - & $>$600 & $>$140k & \checkmark & - & 
  - & - & \checkmark & - & \checkmark & \checkmark & \checkmark & - \\  

  BUFF~\cite{zhang2017detailed} & 
  6 & 3 & $>$30 & $>$13.6k & \checkmark & - & 
  \checkmark & \checkmark & \checkmark & - & \checkmark & \checkmark & \checkmark & \checkmark \\

  DFAUST~\cite{bogo2017dynamic} & 
  10 & $>$10 & $>$100 & $>$40k & \checkmark & - & 
  \checkmark & \checkmark & \checkmark & \checkmark & \checkmark & \checkmark & \checkmark & \checkmark \\  

  People Snapshot~\cite{alldieck2018video} & 
  9 & - & 24 & 15k & \checkmark & - & 
  \checkmark & - & - & - & \checkmark & \checkmark & \checkmark & \checkmark \\

  LiveCap~\cite{habermann2019livecap} & 
  7 & 11 & 11 & 36k & \checkmark & - & 
  \checkmark & - & \checkmark & \checkmark & \checkmark & \checkmark & \checkmark & \checkmark \\

  DynaCap~\cite{habermann2021real} & 
  4 & 5 & 5 & 35k & \checkmark & - & 
  \checkmark & - & \checkmark & \checkmark & \checkmark & \checkmark & \checkmark & \checkmark \\

  DeepCap~\cite{habermann2020deepcap} & 
  4 & 17 & 17 & 26k & \checkmark & \checkmark & 
  \checkmark & - & \checkmark & \checkmark & \checkmark & - & \checkmark & \checkmark \\

  HUMBI~\cite{Yu2020HUMBIAL} & 
  772 & - & - & $\sim$26M & \checkmark & - & 
  \checkmark & - & - & \checkmark & \checkmark & \checkmark & \checkmark & \checkmark \\

  THuman~\cite{Zheng2019DeepHuman} & 
  200 & - & - & $>$6k & - & - & 
  \checkmark & \checkmark & - & - & - & \checkmark & \checkmark & \checkmark \\

  THuman2.0~\cite{tao2021function4d} & 
  200 & - & - & $>$500 & - & - & 
  - & - & - & - & - & \checkmark & \checkmark & \checkmark \\

  \midrule
  \multicolumn{15}{c}{Multi-task} \\
  \midrule
  
  \textbf{\name (ours)}  & \numsubjects & \numactions & \numsequences & \numframes & \checkmark & \checkmark & \checkmark & \checkmark & \checkmark & \checkmark & \checkmark & \checkmark & \checkmark & \checkmark \\
  
  \bottomrule

\end{tabular}
}

\end{center}
\bigskip\centering
\end{table}

\end{document}